%% file: main.tex
\PassOptionsToPackage{svgnames,table}{xcolor}

\documentclass[11pt,letterpaper]{mystyle}

\usepackage[comma,authoryear,compress]{natbib}
\usepackage{xcolor}
\usepackage{hyperref}
\usepackage[all]{hypcap}
\usepackage{cleveref}

\hypersetup{
    colorlinks=true,
    citecolor=YaleBlue,
}

\usepackage{algorithm}
\usepackage{algorithmicx}
\usepackage{algpseudocode}

\usepackage{amsmath}
\usepackage{amssymb}
\usepackage{amsfonts}
\usepackage{mathtools}
\usepackage{amsthm}
\usepackage{mathrsfs}
\usepackage{nicefrac}
\usepackage{dsfont}
\usepackage{fdsymbol}

\usepackage{graphicx}
\usepackage{subcaption}
\usepackage{float}
\usepackage{wrapfig}
\usepackage{rotating}
\usepackage{adjustbox}
\usepackage{capt-of}

\usepackage{booktabs}
\usepackage{bigstrut}
\usepackage{threeparttable}
\usepackage{multirow}
\usepackage{tabularx}
\usepackage{array}
\usepackage{makecell}

\usepackage[font=small,labelfont=bf]{caption}

\usepackage[utf8]{inputenc}
\usepackage[T1]{fontenc}
\usepackage{url}
\usepackage{microtype}
\usepackage{enumitem}
\usepackage{lipsum}
\usepackage{stackengine}
\usepackage{bxcoloremoji}

\input{macro}

\newtcolorbox{AIbox}[2][]{aibox,title=#2,#1}

\definecolor{lightblue}{rgb}{0.22,0.45,0.70}
\definecolor{Gray}{gray}{0.95}
\definecolor{Cornsilk}{rgb}{1.0,0.97,0.86}

\definecolor{setablue}{RGB}{220,232,246}
\definecolor{tmaxgreen}{RGB}{224,239,220}
\definecolor{blockgray}{RGB}{245,245,245}
\definecolor{bestbg}{RGB}{198,224,255}
\definecolor{secondbg}{RGB}{255,222,173}

\makeatletter
\renewcommand{\AB@affilsepx}{\protect\\\protect\Affilfont}
\makeatother

\newcommand{\second}[1]{\underline{#1}}

\title{Credit Where It Matters: Dependency-Aware Policy Optimization for Terminal Agents}

\author{
Yu Li$^1$, Guangfeng Cai$^1$, Long-Fei Li$^2$, Shuo Han$^2$, Shengtian Yang$^1$, \\ 
Han Luo$^1$, Kaibing Yang$^1$, and Lei Feng
}

\affil[1]{School of Computer Science and Engineering, 
  Southeast University, Nanjing, China } 
  
\affil[2]{Huawei Noah's Ark Lab, China}

\correspondingauthor{Lei Feng, fenglei@seu.edu.cn}

\begin{document}

\begin{abstract}
{\centering\textbf{Abstract}\par}
\vspace{0.5em}
Terminal-using agents benefit from reinforcement learning (RL) in coding, debugging, and other multi-step terminal tasks. 
In these tasks, later commands often depend on information or intermediate results produced by earlier commands.
However, existing trajectory-level and step-level credit assignment methods do not explicitly trace the read-write dependencies through which commands affect the final outcome. Consequently, training signals could still be assigned to irrelevant operations, weakening learning from relevant steps. In this paper, we propose Dependency-Aware Group Policy Optimization (DepGPO), which uses execution dependencies between commands to guide credit assignment for terminal agents. Specifically, we construct a command dependency graph from execution traces and trace backward from the resources inspected by the task verifier. We then assign credit to relevant writes and their supporting reads along these paths, and use it to redistribute trajectory advantages across steps. 
Extensive comparative experiments and ablation studies demonstrate that DepGPO improves task performance and training stability on complex terminal tasks.

\end{abstract}

\maketitle
\vspace{3mm}

\section{Introduction}

Terminal-using agents \citep{claudecode, codex, gemini,openhands} have become a major application of language models for coding, debugging, and other multi-step terminal tasks \citep{sweagent, terminal, tmax, endless}. These tasks involve multi-turn interactions with a persistent environment, where later commands use information and intermediate results produced by earlier ones. Recent work improves these agents through reinforcement learning (RL), with outcome rewards determined by task verifiers that check whether the resulting environment satisfies task requirements \citep{seta, tmax, river}. A common approach uses group-based algorithms such as GRPO \citep{deepseekmath} and DAPO \citep{dapo}, which compare outcome rewards across sampled trajectories. In their outcome-based formulations, a trajectory-level advantage is shared across all generated steps. However, commands can play different roles: an inspection may guide a later repair, whereas a write may be overwritten before its output is used. This mismatch calls for step-level credit assignment that reflects how individual commands contribute to task completion.

To refine this within-trajectory credit assignment, recent work has developed several step-level methods. GiGPO compares actions taken from the same environment state \citep{gigpo}, and HGPO further accounts for historical context when forming groups \citep{hgpo}. GAGPO uses grouped state values to propagate outcome supervision through time \citep{gagpo}. Graph-based methods, including GEPO \citep{gepo}, GraphGPO \citep{graphgpo}, and G2PO \citep{g2po}, construct state-transition graphs and derive credit from structural importance, goal progress, or state values. HCAPO provides another route by using a language model to reassess intermediate actions in hindsight \citep{hcapo}. Although these methods enable finer-grained credit assignment, they do not explicitly trace how information and intermediate results flow between commands. In fact, terminal execution traces can expose these read-write dependencies, linking individual commands to the resources checked by the verifier and providing direct evidence for credit assignment.

\begin{figure}[t]
\begin{center}
    \includegraphics[width=\linewidth]{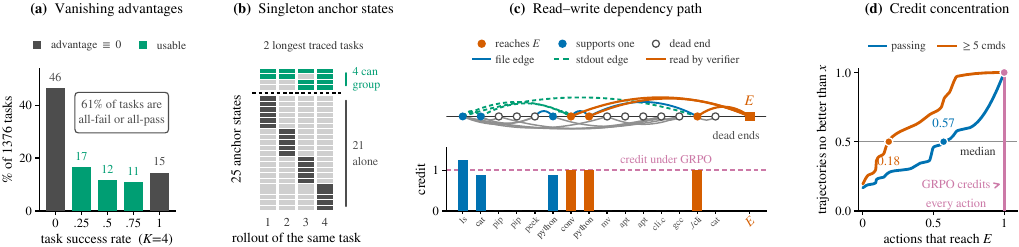}
\end{center}
\caption{
\textbf{Pilot analysis of credit assignment in terminal tasks.}
\textbf{(a)} With $K=4$, an SFT policy yields zero group-relative advantages on $61\%$ of tasks.
\textbf{(b)} Sparse anchor-state recurrence in our terminal implementation of GiGPO.
\textbf{(c)} Example command dependencies and raw credits; $E$ denotes the resources read by the verifier.
\textbf{(d)} Cumulative distributions of the fraction of commands connected to $E$ in passing Terminal-Bench 2.0 reference trajectories ($n=51$) and the subset with at least five commands ($n=18$).
Pink guides illustrate GRPO's uniform step weights.
}
\label{fig:credit-geometry}
  \vskip -0.6cm
\end{figure}

To examine the limitations of existing credit assignment and assess whether execution dependencies can support finer-grained attribution, we conduct a pilot study (Figure \ref{fig:credit-geometry}). With four rollouts per task from a model trained by supervised fine-tuning, 61\% of 1,376 task groups yield identical binary rewards and zero group-relative advantages (Figure 1(a)). DAPO can filter these groups through dynamic sampling, but allocating the remaining nonzero advantages across commands remains unresolved. For GiGPO, exact matching in our terminal implementation yields only four recurring anchor states out of 25 in the two traced tasks with the most anchors, limiting step-level comparisons (Figure 1(b)). We further analyze Terminal-Bench 2.0 reference solutions and denote the set of files and other resources read by the task verifier by \(E\). Execution traces reveal command dependency paths leading to this resource set (Figure 1(c)). Across 51 passing reference trajectories, the median fraction of commands on these paths is 0.57, falling to 0.18 for the 18 trajectories with at least five commands (Figure 1(d)). Together, these findings highlight the need to distinguish command contributions more carefully and show that execution dependencies provide observable evidence for allocating outcome-based learning signals to relevant steps.

Building on these pilot findings, we develop \textbf{Dep}endency-Aware \textbf{G}roup \textbf{P}olicy \textbf{O}ptimization (DepGPO), which redistributes trajectory advantages using command dependencies connected to task verification. Specifically, we construct a directed command graph from recorded resource reads and writes and subsequent uses of command outputs, then trace backward from the verifier's read set \(E\). For file writes, credit is the fraction of written lines that reach $E$. A line reaches $E$ if it remains in a file read by the verifier, or if a later command on a dependency path to $E$ reads it before it is overwritten. Reads then accumulate credit from the relevant writes they support, with each contribution decaying with shortest-path distance in the dependency graph. We then aggregate command credits within each agent step and normalize them into nonnegative, unit-mean step factors. Multiplying these factors by the original trajectory advantage produces per-step advantages for the clipped policy objective. In this design, group-relative outcomes determine the sign of the learning signal, while execution dependencies determine its allocation across steps.

Our main contributions are as follows:

\begin{itemize}[leftmargin=0.5cm,topsep=0pt]
    \item We propose Dependency-Aware Group Policy Optimization (DepGPO), which uses execution dependencies between commands to redistribute group-relative advantages through distinct credit assignment rules for relevant writes and their supporting reads.
    \item We analyze training dynamics through controlled experiments and show that DepGPO delays training degradation and improves the stability of reinforcement learning for terminal agents.
    \item We conduct experiments with Qwen3.5-9B and Qwen3.6-27B using both SETA and TMAX training data, demonstrating consistent improvements in task performance and training stability.
\end{itemize}

\section{Related Work}

\noindent\textbf{Reinforcement Learning with Outcome Supervision.}
Outcome-supervised RL uses response- or trajectory-level rewards as the training signal~\citep{coderl,bcoder}.
PPO estimates advantages with a learned value model~\citep{ppo}, while GRPO uses group-relative rewards without a separate critic~\citep{deepseekmath}.
Such methods have been applied to terminal agents~\citep{t1,river}, including Endless Terminals with PPO~\citep{endless} and SETA with GRPO, task synthesis, and environment evolution~\citep{seta}.
However, sparse terminal rewards provide limited guidance for intermediate commands, and continued policy updates can become unstable~\citep{dapo, liang2026granularity, jin2026revisiting}.

\noindent\textbf{Training Stability.}
Methods for stabilizing RL maintain exploration, constrain policy updates, or reduce rollout--training mismatch.
DAPO combines asymmetric clipping with dynamic sampling~\citep{dapo}, while DPPO directly constrains policy divergence~\citep{dppo}.
For terminal agents, TMAX further combines DPPO with an FP32 language-model head and larger rollout groups to reduce numerical mismatch and training collapse~\citep{tmax}.
These techniques improve optimization stability but do not identify which intermediate commands contribute to the outcome.

\noindent\textbf{Process Feedback and Step-Level Credit Assignment.}
Fine-grained supervision evaluates intermediate behavior through process rewards or step-level comparisons~\citep{credit_assignment_survey, dihan2026swe, han2026swe, li2026phgpo, cai2026beyond, yang2026srpo,yang, yang2026sca, phase}.
Process reward models can use human step labels~\citep{LightmanKBEBLLS24}, sampled continuations~\citep{WangLSXDLCWS24}, or outcome-derived implicit rewards~\citep{prime}; AgentPRM extends process evaluation to multi-turn agents~\citep{agentprm}.
In terminal tasks, River uses rule-based behavioral penalties~\citep{river}, while T1 combines verifier-based rewards with value-model credit assignment~\citep{t1}.
GiGPO instead compares actions from shared environment states~\citep{gigpo}, while GraphGPO~\citep{graphgpo} and G2PO~\citep{g2po} derive step credit from state-transition graphs.
These approaches provide finer feedback, but learned scores require supervision, predefined penalties cover behaviors, and state-based methods depend on representations that may omit file or process changes.

\noindent\textbf{Execution Traces for Terminal Agents.}
Execution traces expose agent actions and their effects on the environment and have been studied for observability, failure attribution, and execution provenance~\citep{agenttrace,agentracer,traceprovenance}.
CodeTracer reconstructs hierarchical state transitions and localizes failure onset for diagnostic replay~\citep{codetracer}, while Shepherd supports branching from recorded intermediate states and computes advantages from additional rollouts~\citep{shepherd}.
DepGPO instead uses the execution trace itself as credit-assignment evidence: it follows read--write dependencies from verifier-inspected resources to identify relevant writes and their supporting reads, then redistributes group-relative advantages accordingly.

\section{Method}
\label{sec:method}

\subsection{Problem Setup and Overview}
\label{sec:method_setup}

Let $\pi_\theta$ denote the language model policy of a terminal agent, where $\theta$ represents the model parameters.
Given a task $x$, the policy generates a response containing terminal commands, whose execution yields an environment observation.
A response and its resulting observation form one interaction step.
Together, the $T_i$ steps of the $i$-th task attempt form a trajectory $\tau_i$.
Since a step may contain several commands or none, we use commands as the units of dependency analysis, while policy optimization operates on the tokens generated in each response.

At the end of each trajectory, a task verifier evaluates the resulting environment and returns a reward $R_i$.
We collect $K$ trajectories for the same task using a fixed sampling policy $\pi_{\theta_{\mathrm{old}}}$.
Following GRPO~\citep{deepseekmath}, we compute the group-relative advantage
\begin{gather}
\widehat A_i=\frac{R_i-\mu_x}{\sigma_x+\epsilon},
\label{eq:depgpo_group_advantage}
\end{gather}
where $\mu_x$ and $\sigma_x$ are the mean and population standard deviation
of the $K$ rewards, and $\epsilon>0$ prevents division by zero.
This advantage is positive when the reward exceeds the group mean and negative when it falls below the mean.
Under outcome supervision, GRPO applies the same trajectory advantage to every token generated during that attempt.

DepGPO redistributes this advantage using execution dependencies within each trajectory.
We build a graph whose nodes are commands from that trajectory, then compute credit for relevant writes and their supporting reads.
Credits of commands executed in the same step are added and normalized into step factors.
These factors determine per-step advantages, which are assigned to the tokens generated in their respective steps.
The following sections describe graph construction, credit assignment, and policy optimization.

\begin{figure}[t]
\begin{center}
    \includegraphics[width=\linewidth]{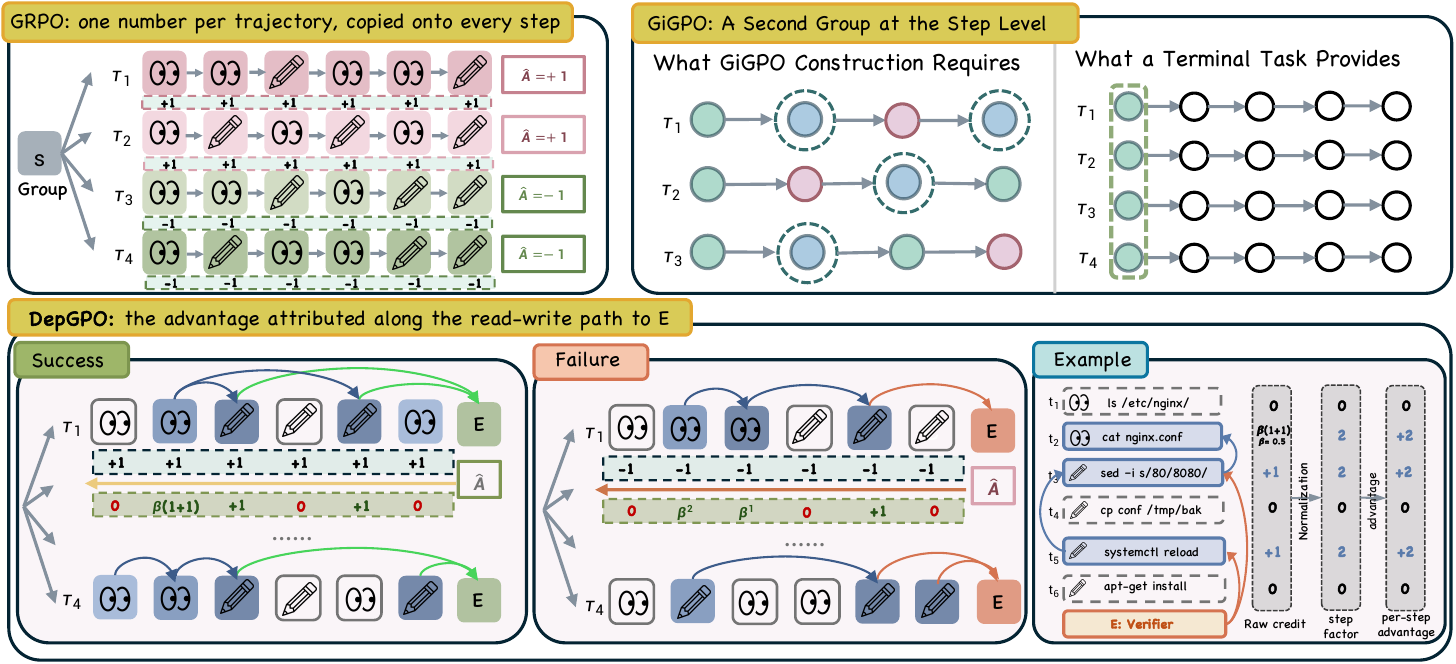}
\end{center}
\caption{
Overview of DepGPO.
Unlike GRPO's uniform trajectory-level assignment and GiGPO's generic step-level grouping, DepGPO traces read--write dependencies to the verifier-inspected resources $E$ and redistributes the group-relative advantage according to dependency-derived step factors.
The example illustrates the transformation from raw credit to the final per-step advantage.
}
\label{fig1}
\vskip -0.6cm
\end{figure}

\subsection{Command Dependency Graph}
\label{sec:method_graph}


We extract file accesses, port and process events, and standard output from command execution traces.
For file writes, we identify written line ranges from differences in file contents before and after execution, when this information is available.
In contrast, file-read records identify the accessed file paths but do not specify which lines are used during computation.
Accordingly, we track writes at line level and record reads at file level.
Based on these records, commands with at least one recorded write are classified as write commands, including those that also read; all remaining commands are classified as read commands.
The two categories are mutually exclusive, and each command receives credit only under the rule for its assigned category.



For each trajectory, we construct a directed command dependency graph $G$ containing both agent commands and the verifier command.
Edges capture dependencies through file contents and standard output and may connect commands across interaction steps.

\noindent\textbf{File edges.}
File edges record the use of content that earlier commands wrote to files.
A file edge $c_u\rightarrow c_v$ means that command $c_v$ reads a file containing lines written by the earlier command $c_u$ that have not yet been overwritten.
We process commands in execution order and record the latest writer of each tracked line.
For the current command, we first connect the earlier writers of these remaining lines to the current command based on its file reads.
We then update the records for lines written by the current command.
An overwrite changes only the records for the replaced lines.
It does not create an edge by itself, because replacing existing content does not establish that the later command read that content.
If the later command also reads the file before replacing its contents, that read creates the corresponding file edge.

\noindent\textbf{Standard-output edges.}
Standard-output edges record the reuse of information printed by commands in the terminal.
To construct these edges, we match typed values, such as file paths and ports, in a later command's text against earlier standard output.
For each match, we add an edge from the command that first printed the value to the command that reuses it.
During this process, we exclude values already supplied in the task description, since their reuse does not establish a dependency on an earlier command.
Because both file and standard-output edges follow execution order, the resulting graph $G$ is acyclic.

We also trace the verifier to identify the resources inspected during evaluation.
We denote these resources by $E$, the verifier's read set.
This set includes files, ports, and processes.
File existence and metadata checks are included, while files written by the verifier itself are excluded from the file set.
These resources provide the starting point for tracing the writes relevant to verification.

\subsection{Dependency-Aware Credit Assignment}
\label{sec:method_credit}

\noindent\textbf{Write credit.}

We assign file-write credit by the fraction of written lines that reach $E$.
A line reaches $E$ directly if its file is inspected by the verifier and the line has not been overwritten by a later agent command.
Otherwise, it may contribute indirectly if a later write command reads the file before the line is overwritten and produces an output that reaches $E$, either directly or through further writes.

We apply a conservative rule to indirect contributions: the consuming command must write to a file other than the input file and must not overwrite the input lines being counted.
The first condition excludes same-file edits, which produce no output elsewhere; earlier lines can still receive direct credit if they remain in a verifier-inspected file.
The second excludes lines overwritten by the consumer from downstream contribution.

\begin{wrapfigure}{r}{0.48\textwidth}
    \vspace{-8pt}
    \centering
    \includegraphics[width=\linewidth]{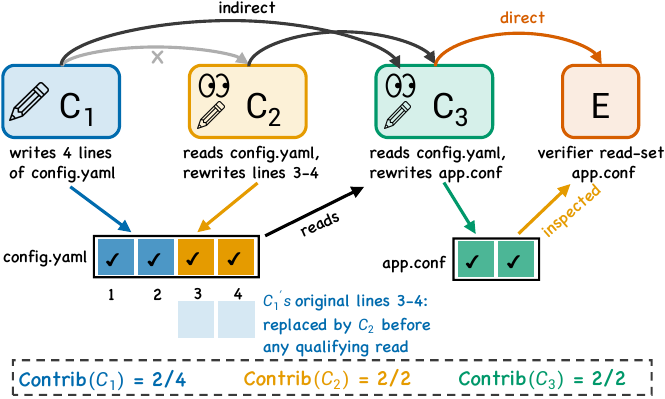}
    \caption{
    Illustration of write-credit assignment through direct and indirect dependencies.
    }
    \label{fig:depgpo_write_credit}
    \vspace{-8pt}
\end{wrapfigure}

Credit assignment proceeds backward from $E$, starting with writes whose contents remain in inspected resources and therefore receive direct credit.
For each identified write, we trace its file reads to earlier written lines satisfying the indirect-contribution conditions, and recursively repeat this process for their writers until no additional writers are found.
These relations form a separate reachability graph $G_E$, whose nodes are agent commands and an endpoint representing $E$, excluding the verifier command.
Complete line-level rules are provided in Appendix~\ref{app:depgpo_indirect}.

Let $\mathcal W_c$ be the set of tracked lines written by command $c$, identified by file and line position, and let $\mathcal W_c^E\subseteq\mathcal W_c$ contain those that contribute to $E$ directly or indirectly.
For $|\mathcal W_c|>0$, the raw credit is
\begin{gather}
q(c)=\operatorname{Contrib}(c)
=\frac{|\mathcal W_c^E|}{|\mathcal W_c|},
\label{eq:depgpo_write_credit}
\end{gather}
where $|\cdot|$ denotes set cardinality.
Each qualifying line contributes one count regardless of the number of downstream consumers or dependency length; downstream contribution scores are not multiplied along the chain.
Thus, $\operatorname{Contrib}(c)\in[0,1]$ measures the fraction of lines written by $c$ that contribute to $E$, without distance decay.

Ports and processes lack a meaningful line-level representation and are therefore handled separately.
A command represented only by such resources receives credit one if its opened port or created process is inspected by the verifier, and zero otherwise.
Resource-specific cases, unavailable line ranges, and commands involving multiple resource types are detailed in the Appendix~\ref{app:depgpo_credit}.

\noindent\textbf{Read credit.}

\begin{wrapfigure}{r}{0.48\textwidth}
    \vspace{-8pt}
    \centering
    \includegraphics[width=\linewidth]{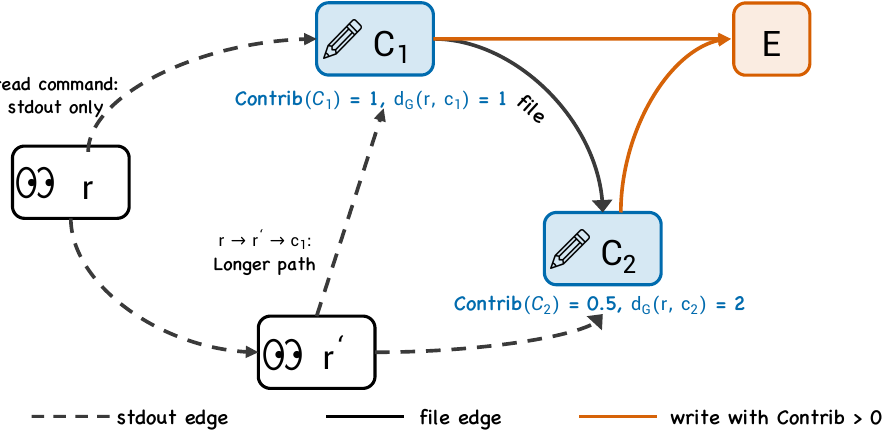}
    \caption{
    Illustration of read-credit assignment through dependency paths in $G$.
    }
    \label{fig:depgpo_read_credit}
    \vspace{-8pt}
\end{wrapfigure}

Read commands can support relevant writes by revealing information used later.
Because a read command produces no recorded file write, its support begins with a standard-output edge; once its output is reused, the path may continue through file or standard-output edges in $G$.
We use these paths to identify the writes supported by each read command.

For a read command $r$, let $\mathcal D_G(r)$ denote the agent write commands reachable from $r$ in $G$, and let $d_G(r,c)$ be the shortest-path length to a reachable write $c$.
Its raw credit is
\begin{gather}
q(r)=\sum\nolimits_{c\in\mathcal D_G(r)}
\beta^{d_G(r,c)}\operatorname{Contrib}(c),
\label{eq:depgpo_read_credit}
\end{gather}
where $\beta\in(0,1]$ controls decay with dependency distance, measured in graph edges rather than interaction steps.
If multiple paths connect $r$ to the same write $c$, its contribution is counted once using the shortest path.
Contributions from distinct positive-credit writes are instead summed, while a read with no path to any positive-credit write receives zero.

Both read and write credits are nonnegative scores of relevance to verification rather than trajectory success.
They determine where the learning signal is assigned, while the group-relative advantage determines whether responses are reinforced or penalized during policy optimization.

\subsection{Advantage Redistribution and Policy Optimization}
\label{sec:method_optimization}

\noindent\textbf{From operation credit to step factors.}
The preceding rules assign raw credit to traced write and read operations.
We aggregate all credits within the same interaction step:
\begin{gather}
q_{i,t}=
\sum_{c\in\mathcal W_{i,t}} q(c)
+
\sum_{r\in\mathcal R_{i,t}} q(r),
\label{eq:depgpo_step_credit}
\end{gather}
where $\mathcal W_{i,t}$ and $\mathcal R_{i,t}$ are the traced write and read operations in step $t$ of trajectory $i$.
Let $\mathcal S_i$ contain the steps with at least one traced operation.
Step credits are compared only within the same trajectory, so the resulting weights reflect relative relevance rather than absolute credit scale.

We use the minimum step credit as a within-trajectory reference and normalize each excess above it by the mean excess:
\begin{gather}
m_i=\min_{s\in\mathcal S_i}q_{i,s},
\qquad
f_{i,t}=
\frac{q_{i,t}-m_i}
{\frac{1}{|\mathcal S_i|}
\sum\nolimits_{s\in\mathcal S_i}(q_{i,s}-m_i)},
\quad t\in\mathcal S_i,
\label{eq:depgpo_step_factor_main}
\end{gather}
where $m_i$ is the minimum credit and $|\mathcal S_i|$ is the number of compared steps.
Thus, $f_{i,t}$ measures relative relevance within trajectory $i$: minimum-credit steps receive zero, larger credits receive proportionally larger factors, and the mean factor over $\mathcal S_i$ is one.

Steps without traced operations are excluded from this comparison and retain the default factor of one.
If all compared credits are identical, the denominator in Eq.~\ref{eq:depgpo_step_factor_main} is zero; if tracing yields no usable credit, no relative comparison is available.
In either case, all step factors are set to one, recovering the original trajectory-level advantage assignment.

\noindent\textbf{From step factors to per-step advantages.}
The step factors redistribute the trajectory-level advantage $\widehat A_i$ across interaction steps.
Directly using $f_{i,t}\widehat A_i$, however, would give longer responses greater influence because the policy loss is averaged over generated tokens.
We therefore normalize the factors by their token-weighted mean.

Let $L_{i,t}$ be the number of generated tokens from step $t$ included in the loss, excluding environment observations.
For a trajectory with $T_i$ steps, we define the token-weighted mean factor as
\begin{gather}
Z_i=
\frac{\sum\nolimits_{t=1}^{T_i}L_{i,t}f_{i,t}}
{\sum\nolimits_{t=1}^{T_i}L_{i,t}},
\end{gather}
and use it to normalize the step factors when defining the final per-step advantage:
\begin{gather}
\widehat A^{\mathrm{Dep}}_{i,t}=
\frac{f_{i,t}}{Z_i}\widehat A_i.
\label{eq:depgpo_token_advantage}
\end{gather}
Because the same $Z_i$ scales every step, this preserves their relative weights while making the token-weighted mean factor one.
Thus, DepGPO redistributes the original trajectory advantage across tokens without changing its overall scale.

Each generated token in step $t$ receives $\widehat A^{\mathrm{Dep}}_{i,t}$.
Larger factors strengthen the learning signal, whereas a zero factor removes that step's direct contribution to the policy loss.
Since all factors are nonnegative, the redistribution preserves the sign of $\widehat A_i$, and a zero advantage remains zero.
If all steps contribute the same number of loss tokens, $Z_i=1$ and $\widehat A^{\mathrm{Dep}}_{i,t}=f_{i,t}\widehat A_i$.

\noindent\textbf{Policy optimization.}
DepGPO retains the GRPO policy optimization procedure and changes only how the group-relative advantage is distributed across interaction steps.
Index the policy-generated tokens in a training minibatch by $n=1,\ldots,N$.
Let $y_n$ denote token $n$ and $h_n$ its context, including the task, preceding observations, and previously generated tokens.
If token $n$ belongs to step $t$ of trajectory $i$, it receives
$a_n=\widehat A^{\mathrm{Dep}}_{i,t}$.
The corresponding probability ratio is
\[
\rho_n(\theta)
=
\frac{\pi_\theta(y_n\mid h_n)}
{\pi_{\theta_{\mathrm{old}}}(y_n\mid h_n)},
\]
where $\pi_{\theta_{\mathrm{old}}}$ is the fixed rollout policy and $\pi_\theta$ is the policy being optimized.

Following the clipped surrogate update of the GRPO baseline, we define
\begin{gather}
\ell_{\mathrm{clip}}(\rho,a)=
\min\!\left\{
\rho a,\,
\operatorname{clip}(\rho,1-\epsilon_{\mathrm{lo}},1+\epsilon_{\mathrm{hi}})a
\right\},
\label{eq:depgpo_clip_main}
\end{gather}
where $\epsilon_{\mathrm{lo}}$ and $\epsilon_{\mathrm{hi}}$ specify the lower and upper clipping ranges.
The clipping limits policy changes once the ratio moves beyond the allowed range in the direction favored by the advantage.

For negative advantages, the baseline further applies dual clipping, lower-bounding the surrogate objective by $\eta a$ with $\eta>1$ to prevent excessively negative values from large probability ratios.
Retaining this update rule, we optimize
\begin{gather}
\mathcal L_{\mathrm{Dep}}(\theta)=
-\frac{1}{N}\sum\nolimits_{n=1}^{N}
\begin{cases}
\max\!\left\{
\ell_{\mathrm{clip}}(\rho_n(\theta),a_n),\,
\eta a_n
\right\},
& a_n<0,\\
\ell_{\mathrm{clip}}(\rho_n(\theta),a_n),
& a_n\geq0.
\end{cases}
\label{eq:depgpo_objective_main}
\end{gather}
The minus sign converts surrogate-objective maximization into loss minimization, and $1/N$ averages over generated tokens.
Environment observations remain in the context but contribute no loss terms.

During each update, the dependency-derived factors and per-step advantages are held fixed.
Positive advantages reinforce sampled tokens and negative advantages penalize them, with dependency factors controlling the signal magnitude across steps.
Because these factors are nonnegative, they change the magnitude but not the sign of the original GRPO signal.
The updated policy then collects new trajectories, after which group-relative advantages and dependency credits are recomputed.

Thus, DepGPO preserves the GRPO objective and clipping configuration while replacing uniform trajectory-level advantage assignment with dependency-aware step-level redistribution.
Token mapping and numerical fallback details are provided in Appendix~\ref{app:depgpo_training}.

\section{Experiments}
\label{sec:depgpo_experiments}

\subsection{Experimental Setup}
\label{sec:depgpo_setup}

\noindent\textbf{Models and datasets.}
We evaluate DepGPO with Qwen3.5-9B~\citep{qwen35} and Qwen3.6-27B~\citep{qwen36_27b}, using SETA~\citep{seta} and TMAX~\citep{tmax} separately, for four model--data configurations.
We evaluate the resulting policies on Terminal-Bench 2.0 and 2.1~\citep{terminal} under the same protocol within each configuration.

\noindent\textbf{Training pipeline and task selection.}
For each configuration, we first perform SFT on the corresponding terminal trajectories and use the resulting checkpoint to select tasks for RL.
Specifically, we sample four attempts per candidate task and retain tasks with both successful and failed outcomes, excluding all-pass and all-fail groups whose group-relative advantages are zero under binary rewards.
The retained pool is then split into RL training and validation sets and fixed across all methods within that configuration.
Thus, methods are compared from the same SFT checkpoint and on the same selected task distribution.

\noindent\textbf{Baselines.}
We compare DepGPO with GRPO~\citep{deepseekmath}, DAPO~\citep{dapo}, and DPPO~\citep{dppo}, together with GiGPO~\citep{gigpo} and GraphGPO~\citep{graphgpo} for step-level credit assignment.
For the controlled comparison with GRPO, DepGPO keeps the initial checkpoint, selected task pool, policy objective, and optimization settings fixed, modifying only the redistribution of group-relative advantages across steps.
Other baselines retain their respective method-specific components and optimization procedures.

\noindent\textbf{Training and evaluation.}
For GRPO and DepGPO, each training batch contains four tasks with $K=8$ trajectories per task, giving 32 trajectories in total.
We use a learning rate of $10^{-6}$, limit each trajectory to 10 interaction steps for SETA and 56 for TMAX and 24,576 generated response tokens, and set the DepGPO dependency-decay coefficient to $\beta=0.5$.
During evaluation, we allow up to 56 interaction steps and estimate pass@1 from five independent attempts per task.
We repeat the full evaluation three times and report mean $\pm$ SD of the overall pass@1.
Complete SFT configurations, task-selection counts and splits and method-specific settings are provided in Appendix~\ref{app:depgpo_experimental_details}.

\definecolor{setablue}{RGB}{220,232,246}
\definecolor{tmaxgreen}{RGB}{224,239,220}
\definecolor{oursbg}{RGB}{240,247,250}
\definecolor{bestbg}{RGB}{198,224,255}
\definecolor{secondbg}{RGB}{255,222,173}
\newcommand{\meanstd}[2]{#1{\tiny$\pm$#2}}

\begin{table}[!t]
\centering
\caption{
Task performance on Terminal-Bench 2.0 and 2.1.
Results are reported as pass@1 (\%), mean $\pm$ SD over three evaluation rounds; blue shading and underlining denote the best and second-best results within each setting, respectively.
}
\label{tab:depgpo_main_results}
\small
\setlength{\tabcolsep}{3.8pt}
\renewcommand{\arraystretch}{1.05}

\begin{threeparttable}
\begin{tabular*}{\textwidth}{
@{\extracolsep{\fill}}
l
cccc
@{\hspace{8pt}}
cccc
@{}
}
\toprule
\multirow{3}{*}{\textbf{Method}}
&
\multicolumn{4}{c}{\textbf{Qwen3.5-9B}}
&
\multicolumn{4}{c}{\textbf{Qwen3.6-27B}}
\\
\cmidrule(lr){2-5}
\cmidrule(lr){6-9}
&
\multicolumn{2}{c}{\cellcolor{setablue}\textbf{SETA}}
&
\multicolumn{2}{c}{\cellcolor{tmaxgreen}\textbf{TMAX}}
&
\multicolumn{2}{c}{\cellcolor{setablue}\textbf{SETA}}
&
\multicolumn{2}{c}{\cellcolor{tmaxgreen}\textbf{TMAX}}
\\
\cmidrule(lr){2-3}
\cmidrule(lr){4-5}
\cmidrule(lr){6-7}
\cmidrule(lr){8-9}
&
TB2.0 & TB2.1
& TB2.0 & TB2.1
& TB2.0 & TB2.1
& TB2.0 & TB2.1
\\
\midrule
SFT
& \meanstd{15.28}{0.81}
& \second{\meanstd{16.48}{1.35}}
& \meanstd{14.01}{2.38}
& \meanstd{10.56}{2.82}
& \meanstd{25.92}{0.72}
& \meanstd{26.37}{1.83}
& \meanstd{23.45}{2.97}
& \meanstd{24.27}{0.78}
\\
GRPO
& \meanstd{11.09}{0.26}
& \meanstd{11.01}{0.67}
& \meanstd{14.91}{2.26}
& \meanstd{17.83}{0.57}
& \meanstd{25.09}{1.11}
& \meanstd{23.37}{0.67}
& \meanstd{23.82}{1.95}
& \meanstd{24.94}{0.45}
\\
DAPO
& \meanstd{13.11}{0.13}
& \meanstd{13.26}{0.67}
& \meanstd{15.13}{1.32}
& \meanstd{13.86}{2.36}
& \meanstd{27.34}{2.16}
& \second{\meanstd{28.16}{1.11}}
& \second{\meanstd{26.74}{2.34}}
& \meanstd{25.09}{0.13}
\\
DPPO
& \second{\meanstd{15.36}{0.91}}
& \meanstd{15.81}{0.47}
& \second{\meanstd{17.23}{1.28}}
& \second{\meanstd{18.95}{2.36}}
& \second{\meanstd{28.24}{1.53}}
& \meanstd{28.01}{0.34}
& \meanstd{26.07}{1.35}
& \second{\meanstd{26.67}{1.94}}
\\
GiGPO
& \meanstd{11.24}{0.39}
& \meanstd{11.39}{1.04}
& \meanstd{12.96}{2.38}
& \meanstd{11.31}{2.63}
& \meanstd{16.55}{0.94}
& \meanstd{16.70}{2.70}
& \meanstd{15.96}{1.70}
& \meanstd{18.65}{2.55}
\\
GraphGPO
& \meanstd{10.79}{0.22}
& \meanstd{10.79}{0.39}
& \meanstd{10.56}{0.22}
& \meanstd{11.46}{1.03}
& \meanstd{17.00}{0.91}
& \meanstd{16.63}{0.22}
& \meanstd{18.20}{0.39}
& \meanstd{16.85}{1.70}
\\
\rowcolor{oursbg}
\textbf{DepGPO}
& \meanstd{\textbf{25.24}}{\textbf{1.01}}
& \meanstd{\textbf{24.64}}{\textbf{2.36}}
& \meanstd{\textbf{20.45}}{\textbf{0.98}}
& \meanstd{\textbf{22.17}}{\textbf{1.80}}
& \meanstd{\textbf{37.75}}{\textbf{1.25}}
& \meanstd{\textbf{37.90}}{\textbf{3.97}}
& \meanstd{\textbf{35.06}}{\textbf{1.35}}
& \meanstd{\textbf{36.70}}{\textbf{0.13}}
\\
\bottomrule
\end{tabular*}
\end{threeparttable}
\end{table}

\subsection{Main Results}
\label{sec:main_results}

\noindent\textbf{Overall performance.}
Table~\ref{tab:depgpo_main_results} shows that DepGPO achieves the best pass@1 across all eight settings.
Compared with the strongest competing RL baseline in each setting, the gains range from 3.22 to 10.03 percentage points; relative to GRPO, they range from 4.34 to 14.53 points.
These consistent improvements suggest that explicitly tracing read--write dependencies provides information that is not captured by trajectory-level or state-based credit alone, allowing the same outcome signal to be concentrated on commands more closely connected to task verification.

\noindent\textbf{Scaling across models and training data.}
The gains persist for both Qwen3.5-9B and Qwen3.6-27B and with either SETA or TMAX.
Notably, the 27B model shows large improvements on both training sources, reaching 37.75/37.90 on SETA and 35.06/36.70 on TMAX.
One plausible explanation is that the larger model can make better use of the finer-grained learning signal provided by dependency-aware redistribution, while the smaller model is more sensitive to the underlying training distribution.
Indeed, the 9B gain over the strongest baseline is substantially larger on SETA than on TMAX, suggesting that model capacity and task distribution jointly affect how much benefit can be extracted from step-level credit.

\subsection{Training Stability}
\label{sec:depgpo_stability}

\begin{figure}[t]
\centering
\includegraphics[width=\textwidth]{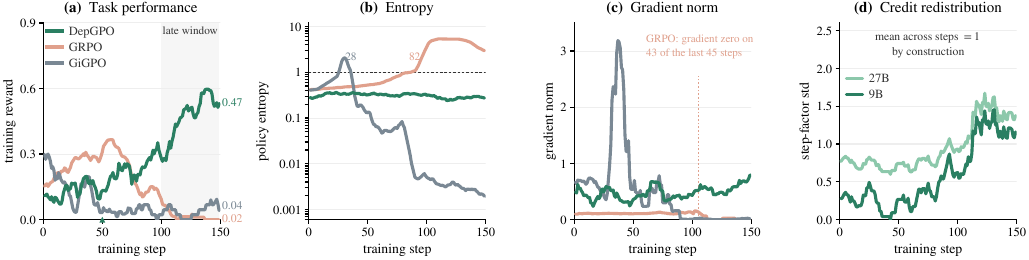}
\caption{
Panels (a)--(c) compare training reward, policy entropy, and gradient norm of DepGPO, GRPO, and GiGPO on Qwen3.5-9B trained with SETA, with shading marking the late training window and entropy plotted on a logarithmic scale.
Panel (d) further shows the standard deviation of DepGPO's step factors across interaction steps, whose mean is fixed at one by normalization.
}
\label{fig:depgpo_stability}
\end{figure}

\noindent\textbf{Reward dynamics.}
Figure~\ref{fig:depgpo_stability}(a) shows that GRPO improves initially but degrades with continued training, while GiGPO deteriorates earlier and remains at a low reward level.
In contrast, DepGPO improves overall and maintains higher rewards in the late training window despite fluctuations.
This difference may reflect their credit-assignment mechanisms: GRPO applies the same group-relative advantage to every step, including commands weakly related to the outcome, whereas GiGPO relies on matching states whose limited recurrence in terminal tasks restricts step-level grouping.

\noindent\textbf{Optimization behavior.}
The entropy and gradient statistics provide further evidence of this divergence.
GiGPO's policy entropy drops sharply after a brief increase, while GRPO's rises substantially later in training; DepGPO remains comparatively stable (Figure~\ref{fig:depgpo_stability}(b)).
Both baselines also approach near-zero gradient norms in later stages, whereas DepGPO retains nonzero gradients (Figure~\ref{fig:depgpo_stability}(c)).
Together with the reward curves, these patterns indicate more stable optimization under DepGPO in the runs shown.

\noindent\textbf{Credit redistribution.}
Figure~\ref{fig:depgpo_stability}(d) reports the standard deviation of DepGPO's step factors, whose mean is fixed at one by normalization.
The standard deviation generally increases during training, indicating increasingly differentiated weights across interaction steps while preserving their average scale.
This pattern suggests that DepGPO becomes increasingly selective in where it allocates the learning signal as training progresses.

\definecolor{fullbg}{RGB}{232,245,233}
\definecolor{basebg}{RGB}{246,246,246}
\definecolor{dropred}{RGB}{178,74,63}
\newcommand{\dropval}[1]{\textcolor{dropred}{\scriptsize$\downarrow$#1}}

\subsection{Ablation Study}
\label{sec:depgpo_ablation}

\noindent
\begin{minipage}[t]{0.38\textwidth}
\vspace{0pt}

We ablate Qwen3.5-9B trained on SETA and evaluate on Terminal-Bench 2.0 and 2.1.
\textbf{Binary write contribution} sets positive $\operatorname{Contrib}(c)$ values to one and leaves zeros unchanged, preserving verifier filtering while removing graded credit.
\textbf{Without verifier filtering} assigns credit one to all write commands, including those that do not reach $E$.
For both variants, read credit is recomputed using the original read-credit rule.

\end{minipage}
\hfill
\begin{minipage}[t]{0.58\textwidth}
\vspace{0pt}
\vspace{-\abovecaptionskip}

\captionof{table}{
Ablation study of Qwen3.5-9B trained on SETA.
Pass@1 (\%), mean $\pm$ SD.
Red arrows denote performance drops relative to Full DepGPO.
}
\label{tab:depgpo_ablation}

\vspace{2pt}

\small
\setlength{\tabcolsep}{3pt}
\renewcommand{\arraystretch}{1.08}

\begin{tabular}{@{}lcc@{}}
\toprule
\textbf{Variant} & \textbf{TB2.0} & \textbf{TB2.1} \\
\midrule

\rowcolor{basebg}
GRPO
& \meanstd{11.09}{0.26}
& \meanstd{11.01}{0.67}
\\

Binary write contribution
& \meanstd{23.07}{0.85}\,\dropval{2.17}
& \meanstd{22.55}{0.91}\,\dropval{2.09}
\\

Without verifier filtering
& \meanstd{18.65}{1.03}\,\dropval{6.59}
& \meanstd{17.75}{1.19}\,\dropval{6.89}
\\

Without supporting-read credit
& \meanstd{21.42}{0.94}\,\dropval{3.82}
& \meanstd{20.15}{1.13}\,\dropval{4.49}
\\

Direct writes only
& \meanstd{23.45}{0.91}\,\dropval{1.79}
& \meanstd{23.00}{1.04}\,\dropval{1.64}
\\

\midrule

\rowcolor{fullbg}
\textbf{Full DepGPO}
& \meanstd{\textbf{25.24}}{\textbf{1.01}}
& \meanstd{\textbf{24.64}}{\textbf{2.36}}
\\

\bottomrule
\end{tabular}

\vspace{2pt}

\end{minipage}

\vspace{0.7em}




\noindent\textbf{Write credit and verifier filtering.}
Binary write contribution reduces performance by 2.17 and 2.09 percentage points on Terminal-Bench 2.0 and 2.1, showing that graded credit provides useful distinctions among relevant writes.
Removing verifier filtering causes a further drop of 4.42 and 4.80 points relative to the binary variant, suggesting that identifying writes connected to verifier-inspected resources is even more important than distinguishing their contribution strength.

\noindent\textbf{Supporting-read credit.}
Removing supporting-read credit lowers performance by 3.82 and 4.49 points.
This drop shows that credit should extend beyond environment-modifying commands: information-gathering reads can guide subsequent relevant writes, and assigning them credit propagates the learning signal to these supporting decisions.

\noindent\textbf{Indirect write contribution.}
Using direct write contributions alone reduces performance by 1.79 and 1.64 points.
This result shows that some useful writes contribute through later write commands even when their own contents do not remain in the resources inspected by the verifier.

Overall, the ablations show that verifier filtering removes unrelated writes, graded credit distinguishes contribution strength, supporting-read credit assigns signal to information-gathering commands, and indirect credit captures earlier writes that support verifier-relevant outputs.

\section{Conclusion}

We introduce DepGPO, which uses execution dependencies to redistribute group-relative advantages across interaction steps in terminal tasks.
Specifically, DepGPO traces written content to the verifier's read set, assigns graded credit to relevant writes, and uses dependency paths to assign credit to their supporting reads.
Experiments on Terminal-Bench 2.0 and 2.1 demonstrate improved task performance, while training analyses show more stable learning.
These findings support observable read-write dependencies as an effective basis for fine-grained credit assignment in terminal agents.

\clearpage
\bibliography{main}

\clearpage

\appendix

\section{Method and Implementation Details}
\label{app:depgpo_details}

DepGPO redistributes the group-relative advantage of a trajectory using execution dependencies that connect agent commands to verifier-inspected resources.
The method traces relevant read and write operations, converts their dependencies into command-level credit, and aggregates these credits into step factors that reweight the advantage over policy-generated tokens.
This appendix details the complete procedure, from dependency construction and verifier-based attribution to credit computation, normalization, policy optimization, fallback handling, and end-to-end training.

\subsection{Notation and Recorded Events}
\label{app:depgpo_notation}

For task $x$, let $\tau_i$ be the $i$-th sampled trajectory and $T_i$ its number of interaction steps. Step $t$ consists of an agent response and the resulting environment observation. The set $\mathcal C_{i,t}$ contains the commands associated with that response, and $\mathcal C_i=\bigcup_t\mathcal C_{i,t}$ contains all agent commands in the trajectory. The verifier command is denoted by $v_i$ and is not a policy-generated action. Its execution provides the terminal reward $R_i$ and the resource-access records used to construct the verifier's read set $E_i$. We omit the trajectory index when discussing one trajectory.

\begin{table}[htbp]
\centering
\small
\caption{Key notation used in the DepGPO credit-assignment procedure.}
\label{tab:depgpo_notation}

\begin{tabularx}{\linewidth}{@{}lX@{}}
\toprule
Symbol & Definition \\
\midrule
$\mathcal C_{\mathrm w},\mathcal C_{\mathrm r}$
& Disjoint sets of agent write commands and read commands. \\
$G,\ G_E$
& Command dependency graph and write-reachability graph. \\
$E$
& Files, ports, and processes inspected by the verifier under the extraction rules. \\
$\mathcal W_c$
& Tracked file records written by command $c$, resolved at line level when available. \\
$D_c,\ I_c$
& Records of $c$ that qualify for direct and indirect contribution. \\
$\operatorname{Contrib}(c)$
& Contribution score of write command $c$. \\
$q(c),\ q(r),\ q_{i,t}$
& Raw write credit, raw read credit, and their aggregate in interaction step $t$. \\
$d_G(r,c),\ \beta$
& Shortest directed dependency distance and its decay coefficient. \\
$\mathcal S_i,\ f_{i,t}$
& Set of steps containing traced commands and the factor assigned to step $t$. \\
$m_i,\ \Delta_i$
& Minimum step credit and total credit above the minimum within trajectory $i$. \\
$L_{i,t},\ Z_i,\ w_{i,t}$
& Included token count, token-weighted normalization factor, and final per-step weight. \\
$\widehat A_i,\ \widehat A^{\mathrm{Dep}}_{i,t}$
& Original group-relative advantage and redistributed advantage for step $t$. \\
$M_{i,j}$
& Loss mask for token position $j$ of trajectory $i$. \\
\bottomrule
\end{tabularx}
\end{table}

\noindent\textbf{Read and write records.}
We record file accesses, port and process events, and standard output. For file writes, differences between the file contents before and after execution identify the written line ranges when available. In contrast, a file-read record identifies the accessed path without identifying which lines the program uses. Accordingly, writes are tracked at line level, while reads are recorded at file level.

A command with at least one recorded write is a write command, including a command that also reads. This rule also applies to recorded deletions and non-file writes. All remaining commands are read commands. Each traced command receives either write credit or read credit according to its command type. The reads of a write command are still used to construct dependency edges and identify indirect contributions of earlier writes.

\noindent\textbf{Record granularity.}
When line ranges are available, each tracked written line is identified by its file path and line position. When line ranges are unavailable, the implementation uses the whole-file marker \texttt{WHOLE\_FILE}$=-1$. In that case, the contribution fraction counts file-level fragments. We distinguish this fallback from line-level tracking when reporting examples and statistics.

\noindent\textbf{Path normalization and resource identity.}
Resource paths are normalized before filtering or matching.
Relative paths are resolved against the tracked working directory, while accesses relative to untracked directory descriptors are discarded rather than inferred.
Agent and verifier records undergo the same normalization and filtering procedure before dependency construction and extraction of $E$.
Resources outside the configured work roots and known runtime artifacts are filtered as system noise.
Symbolic links are not resolved, so resource identity remains path-based and different aliases of the same underlying file are treated as distinct resources.
We regard this as a limitation of the current tracing implementation.

\noindent\textbf{Line ranges and file updates.}
When available, written line ranges are obtained from the new-side hunks of a per-command file diff and recorded in the file coordinates at the time of execution.
Ownership is maintained over $(\text{file},\text{line position})$ pairs and is not rebased after later insertions or deletions.
Consequently, subsequent line shifts may introduce limited positional mismatch in ownership tracking.
Renames are represented as deletion of the source path and a write to the destination path, without transferring ownership between the two paths.

\noindent\textbf{Whole-file fallback.}
When a file-level write is recorded but no usable line range can be recovered, the implementation represents that file with the marker
\texttt{WHOLE\_FILE}$=-1$, which contributes one tracked item to $\mathcal W_c$.
This fallback covers cases such as binary or oversized files, empty-file creation, deletion-only updates, and write sets exceeding the tracing cap.
Accordingly, $\operatorname{Contrib}(c)$ is coarser for such writes than for line-resolved edits, because an unresolved whole-file record contributes a single item to $\mathcal W_c$.

\noindent\textbf{Multiple writes within one command.}
For a command that writes multiple files or disjoint regions, all tracked $(\text{file},\text{position})$ records are pooled into a single set $\mathcal W_c$.
Thus, $|\mathcal W_c|$ is computed across all tracked writes of the command, and unrelated written content can reduce its contribution fraction.
Any command with at least one recorded write is treated as a write command, even if it also performs reads; these reads remain available for dependency construction and for identifying indirect contributions of earlier writers.
Non-file writes such as ports carry no line range and are handled by the binary rules described in Section~\ref{app:depgpo_credit}.

\subsection{Construction of the Command Dependency Graph}
\label{app:depgpo_graph}

The graph $G=(V,\mathcal E_G)$ contains the agent commands and the verifier command, with $V=\mathcal C\cup\{v\}$. A directed edge $c\rightarrow u$ records that a later command $u$ uses information associated with an earlier command $c$. Edges may connect commands from different interaction steps, but never commands from different trajectories.

\noindent\textbf{File dependencies.}
Commands are processed in execution order.
Immediately before command $u$ is processed, let
$o_{u^-}(p,\ell)$ denote the latest recorded writer associated with tracked
line position $\ell$ in file $p$.
File content that has not been written by any agent command has no agent
writer and therefore creates no command dependency.
When $u$ reads a file $p$, we connect every earlier command that still owns
at least one tracked position in $p$ to $u$.
Formally,
\begin{gather}
(c,u)\in\mathcal E_{\mathrm{file}}
\quad\Longleftrightarrow\quad
\exists (p,\ell)\ \text{such that}\ 
p\in\operatorname{ReadFiles}(u),\quad
o_{u^-}(p,\ell)=c.
\label{eq:depgpo_file_edge_app}
\end{gather}
These read dependencies are established before applying the writes of $u$.
The writes then update ownership of the affected positions, while unaffected
positions retain their previous writers.
Consequently, multiple earlier commands may remain represented in the same
file.
An overwrite alone creates no dependency edge between writers; a read of the
earlier content is required.
Thus, the graph captures recorded read-after-write dependencies rather than
write-after-write or write-after-read relations induced solely by execution
order.

This processing order applies to the recorded command as a whole. It does not reconstruct the order of individual instructions within a shell command or its subprocesses.

\noindent\textbf{Standard-output dependencies.}
Standard-output edges capture the reuse of printed information in later command text. The matcher extracts typed values, such as paths and ports, and connects a later command to the first earlier command that printed a matching value of the same type. Values already present in the task description are excluded. This exclusion prevents prompt-provided information from automatically becoming evidence of dependence on an earlier command. When both edge rules connect the same pair of commands, their resource evidence is retained under one mixed edge.

Every edge follows command execution order, making $G$ acyclic. A read command has no recorded write, and therefore has no outgoing file edge under this construction. Its support for later writes begins with a standard-output edge; subsequent edges on the path may be file edges, standard-output edges, or mixed edges.

\subsection{Verifier Read Set and Indirect Write Contributions}
\label{app:depgpo_indirect}
\label{app:depgpo_verifier}

The verifier read set defines the resource endpoints used for attribution.
After the path normalization and filtering described above, let
$\mathcal F_v$ denote the files whose contents are read by the verifier,
$\mathcal B_v$ the files inspected through existence or metadata probes,
and $\mathcal S_v$ the files written by the verifier itself.
The file component of the endpoint set is
\begin{gather}
E_{\mathrm{files}}
=
(\mathcal F_v\setminus\mathcal S_v)
\cup
(\mathcal B_v\setminus\mathcal S_v).
\label{eq:depgpo_E_files}
\end{gather}
Records marked as shadowed by the recorder are removed from the read and probe sets before they are merged.
Port and process checks form the remaining components of $E$ and are not subject to this file-specific filtering.
Thus, a resource may belong to $E$ even when the verifier only probes its existence or metadata rather than reading its contents.

\noindent\textbf{Direct contribution.}
Let $\mathcal W_c$ denote the tracked file records written by command $c$, represented by file and line position when line-level information is available.
Let $o_{\mathrm{end}}(p,\ell)$ denote the writer associated with tracked position $\ell$ in file $p$ at the end of agent execution.
The directly contributing records are
\begin{gather}
D_c=
\{(p,\ell)\in\mathcal W_c:
p\in E_{\mathrm{files}},\
o_{\mathrm{end}}(p,\ell)=c\}.
\label{eq:depgpo_direct_lines}
\end{gather}
A tracked record therefore receives direct credit when it remains attributed to $c$ in a file inspected by the verifier.
For line-resolved records, this corresponds to written lines that remain in the inspected file at the end of agent execution.
The same file-level inspection convention in Eq.~\ref{eq:depgpo_E_files} applies to verifier probes.

\noindent\textbf{Indirect contribution.}
A written record that does not remain in an inspected file may still contribute through a later write command.
For $(p,\ell)\in\mathcal W_c$, let $\mathcal U_c(p,\ell)$ contain later commands $u$ satisfying all of the following conditions:
\begin{enumerate}
\item $u$ is a write command and reads file $p$;
\item immediately before this read is processed, $(p,\ell)$ is still attributed to $c$;
\item $u$ writes to at least one file other than $p$; and
\item the writes of $u$ do not replace the input position $(p,\ell)$ being counted.
\end{enumerate}
These conditions provide a conservative rule for indirect attribution.
The third condition requires information to propagate beyond the input file, while the fourth excludes earlier content that is replaced by the consuming command itself.
Failure to qualify through one command does not preclude direct contribution or indirect contribution through another qualifying downstream write.

We next determine which qualifying consumers ultimately contribute to verification.
For this purpose, we construct a separate reachability graph $G_E$.
It contains agent-command nodes and an endpoint representing $E$, but excludes the verifier command itself.
A qualifying consumption creates an edge from the earlier writer to the later writer, while write commands with directly contributing records are connected to $E$.
Matching non-file resource events provide the corresponding endpoint connections.

Let $\mathcal S_E$ denote the set of write commands that have a directed path to $E$ in $G_E$.
A written record $(p,\ell)$ of command $c$ is therefore counted as an indirect contribution if at least one of its qualifying downstream consumers belongs to $\mathcal S_E$:
\begin{gather}
I_c=
\{(p,\ell)\in\mathcal W_c:
\mathcal U_c(p,\ell)\cap\mathcal S_E\neq\varnothing\},
\qquad
\mathcal W_c^E=D_c\cup I_c.
\label{eq:depgpo_indirect_lines}
\end{gather}
Here, $I_c$ contains the indirectly contributing records of $c$, while
$\mathcal W_c^E$ combines its direct and indirect contributions.
Because this combination is a set union, a record that satisfies both conditions is counted only once.

Operationally, reverse traversal starts from writes connected to resources in $E$ and repeatedly follows qualifying input relations to earlier writers until no additional writers are found.
The input reads of a write command are used only to trace contributions of earlier writes; the command itself remains scored by the write-credit rule.

Each tracked record contributes at most once to $\mathcal W_c^E$, even if it has multiple qualifying consumers or satisfies both the direct and indirect conditions.
An input record is attributed to its earlier writer, whereas a downstream output record belongs to the later writer, and each command is normalized by its own set $\mathcal W_c$.
Qualifying records are not weighted by the contribution scores of downstream writers, and no distance decay is applied in the write-reachability calculation.

\subsection{Complete Command Credit}
\label{app:depgpo_credit}

For a write command with tracked file records, the contribution score is the fraction of its written records that contribute to $E$.
The denominator is the number of tracked records produced by that command, rather than the size of any file it reads or edits.
When line-level ranges are unavailable, \texttt{WHOLE\_FILE}$=-1$ is treated as one tracked record and therefore remains in the file-based branch.
The complete write credit is
\begin{gather}
q(c)=\operatorname{Contrib}(c)=
\begin{cases}
|\mathcal W_c^E|/|\mathcal W_c|, & |\mathcal W_c|>0,\\
1, & |\mathcal W_c|=0\ \text{and}\ c\in\mathcal S_E,\\
0, & \text{otherwise}.
\end{cases}
\label{eq:depgpo_write_complete}
\end{gather}
The second branch applies to write commands with no tracked file record that nevertheless reach $E$, such as commands represented only by port or process events.
For such non-file resources, a match with a verifier-inspected resource gives credit one and no match gives zero.
If a command already has a file-based contribution score, a matching non-file resource does not replace that score, preserving the graded file contribution for commands with mixed resource records.

For a read command $r$, let $\mathcal D_G(r)$ denote the set of agent write commands reachable from $r$ in $G$, excluding the verifier.
Let $d_G(r,c)$ be the shortest directed path length from $r$ to a reachable write $c$.
Its raw credit is
\begin{gather}
q(r)=\sum\nolimits_{c\in\mathcal D_G(r)}
\beta^{d_G(r,c)}\operatorname{Contrib}(c),
\qquad \beta\in(0,1].
\label{eq:depgpo_read_complete}
\end{gather}
Each reachable write contributes at most one term, with the shortest path determining the decay when multiple paths connect $r$ to the same write.
Contributions from distinct writes are summed, so read credit may exceed one.
A read command with no path to any positive-credit write receives zero.

Both write and read credits are nonnegative scores that quantify recorded relevance to verification rather than trajectory success.
They determine where the learning signal is assigned, while the signed group-relative advantage determines whether the corresponding sampled responses are reinforced or penalized.
Thus, high credit identifies greater recorded relevance without implying that the associated content is correct.

\subsection{Advantage Redistribution and Policy Objective}
\label{app:depgpo_training}

\noindent\textbf{From command credits to step factors.}
Each traced command receives either write credit or read credit according to the rules above.
Let $\mathcal C_{i,t}$ contain all traced commands in step $t$ of trajectory $\tau_i$, and let
$\mathcal S_i=\{t:\mathcal C_{i,t}\neq\varnothing\}$ denote the set of steps containing at least one traced command.
We aggregate command credits within each step as
\begin{gather}
q_{i,t}
=
\sum\nolimits_{c\in\mathcal C_{i,t}} q(c),
\label{eq:depgpo_step_credit2}
\end{gather}
where $q(c)$ denotes the write or read credit assigned to command $c$ according to its command type.
Credits from multiple commands in the same step are summed rather than averaged.
The resulting step credits are compared only within the same trajectory.

For nonempty $\mathcal S_i$, define
\begin{gather}
m_i
=
\min_{s\in\mathcal S_i} q_{i,s},
\qquad
\Delta_i
=
\sum\nolimits_{s\in\mathcal S_i}
(q_{i,s}-m_i).
\label{eq:depgpo_shift}
\end{gather}
The step factors are then
\begin{gather}
f_{i,t}
=
\begin{cases}
|\mathcal S_i|(q_{i,t}-m_i)/\Delta_i,
& t\in\mathcal S_i,\ \Delta_i>10^{-9},\\
1,
& \text{otherwise}.
\end{cases}
\label{eq:depgpo_step_factor}
\end{gather}
Thus, when the compared step credits are nondegenerate, subtracting $m_i$ uses the lowest observed credit as the within-trajectory reference, while division by $\Delta_i/|\mathcal S_i|$ normalizes the mean factor over $\mathcal S_i$ to one.
The minimum-credit steps therefore receive factor zero, even when their raw credits are positive.

Steps without traced commands retain an initial factor of one and are not included in the within-trajectory comparison.
If $\mathcal S_i$ is empty, tracing yields no usable credit, or $\Delta_i\leq10^{-9}$, all step factors are set to one and the baseline advantage assignment is retained.
In particular, if no write command has positive contribution to $E$, all write credits are zero and the propagated read credits are also zero, triggering the same fallback.

\noindent\textbf{From step factors to final weights and advantages.}
Let $M_{i,j}\in\{0,1\}$ indicate whether token position $j$ enters the policy loss.
Policy-generated tokens have mask one, while environment observations have mask zero.
The factor of step $t$ is assigned to the corresponding policy-generated token span.
Positions outside the mapped spans retain an initial factor of one; if such positions enter the loss, they are included in the subsequent token-level normalization.

Because the policy loss averages over generated tokens, steps with longer responses contribute more loss terms.
We therefore normalize the step factors by their token-weighted mean.
When all included tokens are covered by mapped step spans, let $L_{i,t}$ denote the number of loss-included tokens from step $t$.
For positive included length and normalization denominator,
\begin{gather}
Z_i=
\frac{\sum\nolimits_{t=1}^{T_i}L_{i,t}f_{i,t}}
{\sum\nolimits_{t=1}^{T_i}L_{i,t}},
\qquad
w_{i,t}=\frac{f_{i,t}}{Z_i},
\qquad
\widehat A^{\mathrm{Dep}}_{i,t}
=
w_{i,t}\widehat A_i.
\label{eq:depgpo_full_token_normalization}
\end{gather}
Here, $f_{i,t}$ is the step factor before token-level normalization, $w_{i,t}$ is the final weight, and $\widehat A^{\mathrm{Dep}}_{i,t}$ is the redistributed advantage for step $t$.
Every loss-included token in that step receives the same redistributed advantage.
A step without traced commands has $f_{i,t}=1$ before token-level normalization and therefore receives final weight $1/Z_i$, which is not necessarily one.

In the implementation, $Z_i$ is computed directly over all policy-generated
tokens included in the loss.
Each token inherits the factor of its mapped interaction step; any included
token without a valid step mapping retains the default factor of one.
When every included token is mapped to a step, this token-level computation
reduces exactly to the expression above.

\noindent\textbf{What normalization preserves.}
For a nondegenerate step-level redistribution,
\begin{gather}
\sum_{t\in\mathcal S_i} f_{i,t}=|\mathcal S_i|.
\end{gather}
Since steps outside $\mathcal S_i$ retain factor one, the mean step factor over the trajectory is also one.
After token-level normalization,
\begin{gather}
\frac{\sum_t L_{i,t}w_{i,t}}
{\sum_t L_{i,t}}=1,
\qquad
\frac{\sum_t L_{i,t}\widehat A^{\mathrm{Dep}}_{i,t}}
{\sum_t L_{i,t}}
=
\widehat A_i.
\label{eq:depgpo_mean_preservation}
\end{gather}
Thus, the step-level normalization preserves the mean step factor, while the token-level normalization preserves the token-weighted mean final weight.
Consequently, the token-weighted mean redistributed advantage remains equal to the original group-relative advantage $\widehat A_i$.

These identities apply to the advantage coefficients rather than to the resulting optimization dynamics.
The policy gradient, its norm, and the clipped objective may still change because different tokens receive different coefficients.
In particular, unequal response lengths can yield $Z_i\neq1$ even when the unweighted mean step factor is one.

All valid weights are nonnegative and therefore preserve the sign of $\widehat A_i$.
If $\widehat A_i=0$, all redistributed advantages remain zero.
DepGPO therefore redistributes the learning signal supplied by the GRPO baseline without creating a nonzero signal when the original group-relative advantage is zero.

\noindent\textbf{Policy objective.}
Index all policy-generated tokens included in a training minibatch by $n=1,\ldots,N$.
Let $y_n$ denote token $n$, $h_n$ its original context, and $a_n$ the redistributed advantage of the interaction step containing that token.
The corresponding probability ratio is
\begin{gather}
\rho_n(\theta)=
\frac{\pi_\theta(y_n\mid h_n)}
{\pi_{\theta_{\mathrm{old}}}(y_n\mid h_n)},
\label{eq:depgpo_ratio}
\end{gather}
where $\pi_{\theta_{\mathrm{old}}}$ is the policy used to collect the trajectories.
This indexing affects only the loss reduction; each token retains the context of its original trajectory.

Following the clipped surrogate update of the GRPO baseline, we define
\begin{align}
\ell_{\mathrm{clip}}(\rho,a)
&=
\min\!\left\{
\rho a,\,
\operatorname{clip}
(\rho,1-\epsilon_{\mathrm{lo}},1+\epsilon_{\mathrm{hi}})a
\right\},
\nonumber\\
\ell_{\mathrm{dc}}(\rho,a)
&=
\begin{cases}
\max\!\left\{
\ell_{\mathrm{clip}}(\rho,a),\eta a
\right\},
& a<0,\\
\ell_{\mathrm{clip}}(\rho,a),
& a\geq0,
\end{cases}
\label{eq:depgpo_clipping}\\
\mathcal L_{\mathrm{Dep}}(\theta)
&=
-\frac{1}{N}
\sum\nolimits_{n=1}^{N}
\ell_{\mathrm{dc}}(\rho_n(\theta),a_n).
\label{eq:depgpo_objective}
\end{align}
Here, $\epsilon_{\mathrm{lo}}$ and $\epsilon_{\mathrm{hi}}$ define the lower and upper clipping ranges, and $\eta>1$ is the dual-clipping threshold for negative advantages.
The dependency-derived factors and redistributed advantages are held fixed during each policy update.
DepGPO changes only the advantages assigned to tokens, while retaining the clipping rule and token-mean reduction of the GRPO baseline.

\noindent\textbf{Fallback and integration.}
Table~\ref{tab:depgpo_fallback} summarizes the conditions under which unit dependency factors are used.
Step factors are derived from the rollout records, expanded to the corresponding token positions, and normalized using the policy-loss mask.
The resulting weights are applied after the baseline group-relative advantages are computed and before the actor update.
The current integration also applies the same weights to the returns tensor.
Because the GRPO path used here is critic-free, this does not introduce a separate critic target; the returns tensor carries the same outcome-derived signal used by the policy update.

\begin{table}[htbp]
\centering
\small
\setlength{\tabcolsep}{4pt}
\renewcommand{\arraystretch}{1.08}
\caption{
Fallback conditions that retain the baseline GRPO advantage assignment.
These cases use unit dependency factors, or disable dependency weighting in the baseline configuration, without removing the trajectory from the baseline reward and group-relative advantage computation.
}
\label{tab:depgpo_fallback}

\begin{tabular}{@{}p{0.49\linewidth}p{0.45\linewidth}@{}}
\toprule
\textbf{Condition} & \textbf{Result} \\
\midrule

No step information or unavailable credit module
& Unit factors. \\

Missing usable verifier or agent-command records
& Unit factors; the missing-trace case is recorded. \\

Empty command-credit output
& Unit factors. \\

$\mathcal S_i=\varnothing$
& Unit factors. \\

$\Delta_i\leq10^{-9}$
& Unit factors, including equal-credit and all-zero cases. \\

Exception caught by the calling trainer
& Unit factors and an error diagnostic. \\

Baseline configuration
& Dependency weighting is disabled. \\

\bottomrule
\end{tabular}
\end{table}

\subsection{End-to-End Procedure}
\label{app:depgpo_algorithm}

The following procedure summarizes the complete DepGPO training pipeline.
Resource extraction and dependency-based credit assignment are computed from the recorded trajectories and are held fixed during the corresponding policy updates.

\begin{enumerate}
\item Sample $K$ trajectories for each task using $\pi_{\theta_{\mathrm{old}}}$, recording step-to-command mappings and policy-generated token spans.

\item Execute the task verifier for each trajectory to obtain the terminal reward $R_i$ and the verifier resource-access records.

\item For each trajectory, extract the verifier endpoint set $E$, construct the command dependency graph $G$, and build the qualifying write-consumption relations that define $G_E$.

\item Traverse $G_E$ backward from $E$ to identify directly and indirectly contributing write records, count each qualifying record once, and compute write credit.

\item For each read command, identify reachable write descendants in $G$ and sum their contribution scores with shortest-path decay to obtain read credit.

\item Aggregate command credits within each interaction step, compute the corresponding step factors, and apply the documented unit-factor fallbacks when necessary.

\item Expand the step factors to policy-generated tokens and normalize them over the policy-loss mask to obtain the final weights.
After the baseline group-relative advantages are computed, apply these weights to obtain the redistributed per-step advantages.

\item Update the policy parameters $\theta$ using Eq.~\ref{eq:depgpo_objective}, and then collect the next group of trajectories with the updated policy.
\end{enumerate}

\section{Experimental Setup and Reproducibility}
\label{app:training_setup}
\label{app:depgpo_experimental_details}

This appendix provides the complete experimental setup used to evaluate DepGPO and the baselines, with particular attention to training consistency, comparison fairness, and reproducibility.
We first describe the common training pipeline and controlled settings shared across methods, followed by the training data, task selection and splits, model configurations, optimization hyperparameters, evaluation protocol, and implementation details.

All methods follow the same two-stage training procedure, consisting of supervised fine-tuning (SFT) on terminal trajectories followed by reinforcement learning (RL) on selected training tasks.
SETA \citep{seta} and TMAX \citep{tmax} are used separately with Qwen3.5-9B \citep{qwen35} and Qwen3.6-27B \citep{qwen36_27b}.
Within each model and data setting, all methods share the same SFT checkpoint, selected task pool, training and validation split, and agent scaffold, ensuring matched starting policies and task distributions.
For the direct comparison between GRPO and DepGPO, we additionally keep the policy objective and optimization settings unchanged, with DepGPO modifying only how the group-relative advantage is redistributed across interaction steps.
Other baselines retain their respective method-specific algorithmic components while using the same shared experimental setting whenever applicable.

\subsection{Supervised Fine-Tuning}
\label{app:sft}
\label{app:depgpo_models_data}

We fine-tune each base model on SETA and TMAX separately, producing an SFT checkpoint for the corresponding RL setting. The SFT configuration uses AdamW with a learning rate of $10^{-5}$, optimizer coefficients $(0.9,0.95)$, weight decay $0.01$, and gradient clipping at $1.0$. The learning rate increases during the first $3\%$ of training and remains constant afterward. Each run completes one epoch over its training corpus.

Training updates all model parameters in bfloat16 and uses Fully Sharded Data Parallel (FSDP), gradient checkpointing, and sequence packing. Multi-turn conversations are tokenized with the preceding context retained, while the loss includes only assistant tokens. Conversations exceeding the sequence limit are truncated from the left. Table~\ref{tab:app_sft} summarizes the corpus sizes and sequence limits. SETA and TMAX use different sequence budgets, while the settings share the optimizer configuration above. The conversation counts refer to SFT examples and are distinct from the task counts used for RL. 

\begin{table}[htbp]
\centering\small
\setlength{\tabcolsep}{3pt}
\renewcommand{\arraystretch}{1.1}
\caption{Supervised fine-tuning configurations. The 9B and 27B columns use Qwen3.5-9B and Qwen3.6-27B, respectively.}
\label{tab:app_sft}
\label{tab:depgpo_run_manifest}
\begin{tabular}{p{0.35\linewidth}cccc}
\toprule
Setting & SETA/9B & SETA/27B & TMAX/9B & TMAX/27B \\
\midrule
Training conversations & $79{,}882$ & $79{,}882$ & $16{,}355$ & ${16{,}355}$ \\
Validation conversations & $3{,}690$ & $3{,}690$ & $166$ & ${166}$ \\
Maximum sequence length & $8{,}192$ & $8{,}192$ & $32{,}768$ & ${32{,}768}$ \\
Per-device micro-batch & $1$ & $1$ & $1$ & ${1}$ \\
Devices & $6$ & $4$ & $4$ & ${4}$ \\
Epochs & $1$ & $1$ & $1$ & ${1}$ \\
Seed & $1$ & $1$ & $1$ & ${1}$ \\
\bottomrule
\end{tabular}
\end{table}

\subsection{Selecting Training Tasks}
\label{app:mining}

With binary outcome rewards, a sampled group containing only successes or only failures has zero centered group-relative advantages. Such groups therefore provide no outcome-based policy-gradient signal for GRPO or for DepGPO's redistribution of those advantages. To increase the frequency of informative groups, we select tasks on which the corresponding SFT policy produces both successful and unsuccessful attempts.

For each candidate task $x$, we sample four trajectories with the SFT checkpoint of that model and data setting. Let $z_{x,k}\in\{0,1\}$ be the verifier outcome of attempt $k$. We compute
\begin{gather}
\widehat p_{\mathrm{SFT}}(x)
=\frac{1}{4}\sum_{k=1}^{4}z_{x,k},
\qquad
\mathcal P
=\{x:0<\widehat p_{\mathrm{SFT}}(x)<1\},
\label{eq:depgpo_task_selection}
\end{gather}
where $\mathcal P$ denotes the retained task pool. With four attempts, this rule retains success rates of $0.25$, $0.50$, and $0.75$. Tasks with zero or four successful attempts are excluded. The retained pool is then fixed across all RL methods in that setting. This selection increases the availability of mixed-outcome tasks under the initial policy, although later training groups can still contain identical rewards.

\begin{table}[htbp]
\centering\small
\setlength{\tabcolsep}{3pt}
\renewcommand{\arraystretch}{1.1}
\caption{Task selection using four attempts per candidate task. Each setting uses its own SFT checkpoint.}
\label{tab:app_mining}
\begin{tabular}{p{0.35\linewidth}cccc}
\toprule
Setting & SETA/9B & SETA/27B & TMAX/9B & TMAX/27B \\
\midrule
Candidate tasks & $1{,}376$ & $1{,}376$ & $2{,}998$ & ${2{,}998}$ \\
Zero successful attempts & $639$ & $279$ & ${2{,}200}$ & ${1{,}950}$ \\
Four successful attempts & $200$ & $501$ & ${258}$ & ${400}$ \\
\textbf{Retained tasks} & $\mathbf{537}$ & $\mathbf{596}$ & $\mathbf{540}$ & $\mathbf{648}$ \\
\midrule
RL training tasks & $483$ & $536$ & $486$ & ${583}$ \\
RL validation tasks & $54$ & $60$ & $54$ & ${65}$ \\
\bottomrule
\end{tabular}
\end{table}

For SETA/9B, the excluded groups contain $639+200=839$ tasks, accounting for approximately $61\%$ of the $1{,}376$ candidates. This is the proportion reported in the pilot analysis. On the same candidate collection, the 27B checkpoint produces fewer zero-success groups and more four-success groups, leaving $596$ tasks for RL. Because selection depends on the initial policy, task pools are constructed separately for each model and data setting. Fairness across methods is maintained by reusing the same pool within a setting.

In the TMAX completion, the 9B counts retain 540 of 2,998 tasks (18.01\%), while the 27B counts retain 648 tasks (21.61\%). The 27B pool contains 583 training tasks and 65 validation tasks after the same deterministic split.

We split each retained pool by sorting task identifiers and assigning the first $\lfloor0.9|\mathcal P|\rfloor$ tasks to RL training and the remainder to RL validation. This deterministic rule gives the counts in Table~\ref{tab:app_mining}. The RL validation tasks are held out from policy updates and are used for checkpoint selection. They are separate from the conversation-level validation splits listed for SFT.

\subsection{Reinforcement Learning}
\label{app:rl}
\label{app:depgpo_protocol}

RL starts from the SFT checkpoint corresponding to the model and training source. Each update samples four training tasks and eight trajectories per task, giving 32 trajectories. The training budget is 150 policy updates. Table~\ref{tab:app_rl} gives the common settings for the controlled GRPO and DepGPO comparison, including the settings that differ across model sizes or training sources.

\begin{table}[htbp]
\centering\small
\setlength{\tabcolsep}{3pt}
\renewcommand{\arraystretch}{1.1}
\caption{RL configurations for the GRPO and DepGPO comparison. Each column starts from its corresponding SFT checkpoint. Method-specific baseline components are retained separately.}
\label{tab:app_rl}
\label{tab:depgpo_train_config}
\begin{tabular}{p{0.35\linewidth}cccc}
\toprule
Setting & SETA/9B & SETA/27B & TMAX/9B & TMAX/27B \\
\midrule
Tasks per update & $4$ & $4$ & $4$ & ${4}$ \\
Trajectories per task $K$ & $8$ & $8$ & $8$ & ${8}$ \\
Trajectories per update & $32$ & $32$ & $32$ & ${32}$ \\
Policy updates & $150$ & $150$ & $150$ & ${150}$ \\
\midrule
Optimizer & AdamW & AdamW & AdamW & AdamW \\
Learning rate & $10^{-6}$ & $10^{-6}$ & $10^{-6}$ & ${10^{-6}}$ \\
Learning-rate schedule & Constant & Constant & Constant & Constant \\
Clip range $(\epsilon_{\mathrm{lo}},\epsilon_{\mathrm{hi}})$ & $(0.2,0.28)$ & $(0.2,0.28)$ & $(0.2,0.28)$ & ${(0.2,0.28)}$ \\
Dual-clip threshold $\eta$ & $3.0$ & $3.0$ & $3.0$ & ${3.0}$ \\
KL penalty / entropy bonus & $0/0$ & $0/0$ & $0/0$ & ${0/0}$ \\
Group-SD normalization & Enabled & Enabled & Enabled & Enabled \\
Response-token budget & $24{,}576$ & $24{,}576$ & $24{,}576$ & ${24{,}576}$ \\
Decay $\beta$ (DepGPO) & $0.5$ & $0.5$ & $0.5$ & ${0.5}$ \\
\midrule
Maximum interaction steps & $10$ & $10$ & $56$ & ${56}$ \\
Sampling temperature & $1.0$ & $1.0$ & $0.7$ & ${0.7}$ \\
Command timeout (seconds) & $120$ & $400$ & $120$ & ${400}$ \\
Sequence-parallel degree & $1$ & $4$ & $1$ & ${4}$ \\
Devices & $4$ & $4$ & $4$ & ${4}$ \\
\bottomrule
\end{tabular}
\end{table}

Across all settings, each update samples four tasks with $K=8$ trajectories per task, yielding 32 trajectories, and training runs for 150 policy updates.
We use AdamW with a learning rate of $10^{-6}$, a constant schedule, clipping range $(0.2,0.28)$, dual-clip threshold $\eta=3.0$, no KL penalty or entropy bonus, group-SD normalization, a response-token budget of 24,576, and $\beta=0.5$ for DepGPO.
The SETA settings allow up to 10 interaction steps per trajectory, while TMAX uses 56, with sampling temperatures of $1.0$ and $0.7$, respectively.
For Qwen3.5-9B, we use a 120-second command timeout and sequence-parallel degree one; Qwen3.6-27B uses a 400-second timeout and sequence-parallel degree four.
All settings use four NVIDIA H200 GPUs.

A trajectory ends when the agent declares completion or reaches its interaction or token limit. The task verifier then returns a binary outcome, with no intermediate environment reward. DepGPO uses the command and verifier records to compute write and read credit, aggregates command credits into step factors, and normalizes those factors into final weights. The weights multiply the group-relative advantages before the policy update, following Appendix~\ref{app:depgpo_training}. The outcome reward and the underlying policy objective remain unchanged in the GRPO comparison.

\subsection{Checkpoint Selection}
\label{app:ckpt}

We periodically evaluate checkpoints on the RL validation tasks from the corresponding retained pool. For each method, we select the checkpoint with the highest validation success rate and use that checkpoint for the final benchmark evaluation. The same selection criterion applies to DepGPO and all baselines. Consequently, a baseline that deteriorates later in training can still be evaluated at an earlier checkpoint when that checkpoint has the highest validation score. The complete training curves are analyzed separately to assess training stability.

Checkpoint selection uses RL validation performance. Terminal-Bench 2.0 and 2.1 scores are reserved for reporting the selected policies, with the data-separation protocol described in Appendix~\ref{app:isolation}.

\subsection{Benchmark Evaluation and Score Aggregation}
\label{app:depgpo_statistics}

We evaluate the selected checkpoints separately on Terminal-Bench 2.0 and 2.1. Each attempt begins from the task's initial environment and allows at most 56 interaction steps. We conduct five independent attempts per task and repeat the complete evaluation three times. Within each round, pass@1 is estimated by averaging the five binary outcomes for each task and then averaging across tasks.

For benchmark version $v$, let $J_v$ be the number of tasks and let $z_{j,h,a}^{(v,m)}\in\{0,1\}$ denote the outcome for method $m$ on task $j$, evaluation round $h$, and attempt $a$. The task and benchmark scores for each round are
\begin{gather}
\widehat p_{j,h}^{(v,m)}=\frac{1}{5}\sum_{a=1}^{5}z_{j,h,a}^{(v,m)},
\qquad
s_h^{(v,m)}=\frac{100}{J_v}\sum_{j=1}^{J_v}\widehat p_{j,h}^{(v,m)}.
\label{eq:depgpo_eval_score}
\end{gather}
With $H=3$ evaluation rounds, we report
\begin{gather}
\overline s^{(v,m)}=\frac{1}{H}\sum_{h=1}^{H}s_h^{(v,m)},
\qquad
\operatorname{SD}^{(v,m)}=
\sqrt{\frac{1}{H-1}\sum_{h=1}^{H}
\left(s_h^{(v,m)}-\overline s^{(v,m)}\right)^2}.
\label{eq:depgpo_eval_sd}
\end{gather}
The table~\ref{tab:depgpo_main_results} show $\overline s\pm\operatorname{SD}$ in percentage points. This SD measures evaluation variability for a fixed trained checkpoint. Variability across independently trained policies requires separate training runs. In addition, averaging individual attempts estimates pass@1; treating a task as solved whenever any of its five attempts succeeds would instead estimate pass@5.

\subsection{Data Separation and Verifier Information}
\label{app:isolation}
\label{app:depgpo_verifier_access}

Task selection uses the training-source collections, and RL validation tasks are excluded from policy updates. Terminal-Bench evaluation tasks are kept separate from the training and validation task pools and are not used to choose checkpoints. Conversation-level SFT splits and task-level RL splits serve different purposes, so their separation is assessed at the appropriate unit.

After a training trajectory is completed, the task verifier returns the terminal outcome and records which resources it inspects.
DepGPO uses these access records to construct $E$ and then computes the dependency-based weights for the finished trajectory.
These weights are applied only after the trajectory has been generated; the policy does not receive $E$ or the verifier access records as additional input during action generation.
Thus, DepGPO still learns from a terminal reward, while execution dependencies are used only to determine how that reward-derived advantage is distributed across steps.
Whether the agent can directly access verifier scripts through the terminal is determined by the agent scaffold and is independent of this weighting procedure.

\section{Task-Level Paired Comparison}
\label{sec:depgpo_task_comparison}

This appendix examines how the aggregate performance gains of DepGPO are distributed across individual tasks.
We compare DepGPO and GRPO using Qwen3.5-9B trained on SETA, with evaluation on Terminal-Bench 2.0 and 2.1.
The analysis examines the frequency and magnitude of improvements, ties, and regressions.

\begin{figure}[ht]
    \centering
    \includegraphics[width=\columnwidth]{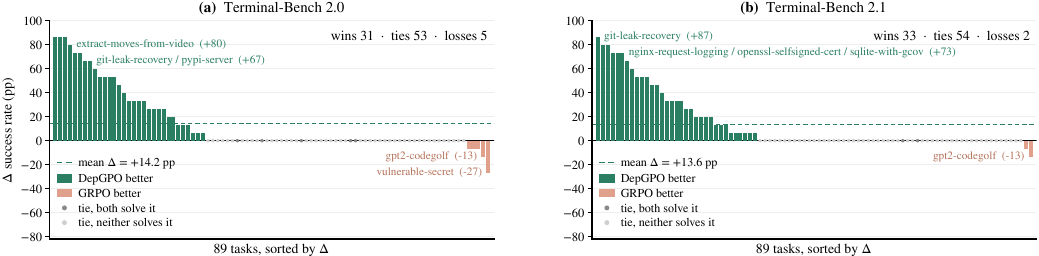}
    \caption{
    Task-level success-rate differences between DepGPO and GRPO using Qwen3.5-9B trained on SETA and evaluated on Terminal-Bench 2.0 (a) and 2.1 (b).
    Each panel independently sorts 89 tasks by $\Delta$, computed as DepGPO minus GRPO in percentage points.
    Green and orange bars indicate improvements and regressions, gray markers denote ties, and dashed lines show the mean difference over all tasks.
    Annotations highlight selected task-level differences.
    }
    \label{fig:depgpo_task_difference}
\end{figure}

\noindent\textbf{Comparison protocol.}
For each benchmark version $v$, we compare the two methods on the same 89 tasks.
Let $\bar p_j^{(v,m)}$ denote the observed success rate of method $m$ on task $j$, expressed as a proportion.
The task-level difference and its benchmark average are
\begin{align}
\Delta_j^{(v)}
&=
100\left(
\bar p_j^{(v,\mathrm{DepGPO})}
-
\bar p_j^{(v,\mathrm{GRPO})}
\right),
\nonumber\\
\overline{\Delta}^{(v)}
&=
\frac{1}{89}\sum_{j=1}^{89}\Delta_j^{(v)}.
\label{eq:depgpo_task_difference}
\end{align}
Positive, zero, and negative differences indicate improvements, ties, and regressions, respectively.
Figure~\ref{fig:depgpo_task_difference} sorts tasks independently within each benchmark by descending $\Delta_j^{(v)}$.
The horizontal axis therefore represents the rank of the observed difference, not task difficulty or a shared task ordering across benchmark versions.

\noindent\textbf{Coverage of improvements.}
On Terminal-Bench 2.0, DepGPO improves on 31 tasks, ties on 53, and performs worse on 5, with a mean improvement of 14.2 percentage points (pp).
On Terminal-Bench 2.1, the corresponding counts are 33, 54, and 2, with a mean improvement of 13.6 pp.
Among tasks with nonzero differences, improvements account for 31 of 36 tasks (86.1\%) on Terminal-Bench 2.0 and 33 of 35 tasks (94.3\%) on Terminal-Bench 2.1.
Thus, when the observed success rates differ, the difference predominantly favors DepGPO.
These counts describe tasks with higher success rates, not necessarily tasks newly solved by DepGPO.

\noindent\textbf{Magnitude of improvements and regressions.}
The positive differences vary substantially across tasks, with several large improvements alongside smaller gains.
On Terminal-Bench 2.0, the highlighted examples include approximately 80 pp on \texttt{extract-moves-from-video} and 67 pp each on \texttt{git-leak-recovery} and \texttt{pypi-server}.
On Terminal-Bench 2.1, \texttt{git-leak-recovery} improves by approximately 87 pp, while \texttt{nginx-request-logging}, \texttt{openssl-selfsigned-cert}, and \texttt{sqlite-with-gcov} each improve by approximately 73 pp.
These examples show that substantial improvements occur on multiple tasks.

Regressions remain present and should be considered alongside these gains.
On Terminal-Bench 2.0, the highlighted decreases are approximately 27 pp on \texttt{vulnerable-secret} and 13 pp on \texttt{gpt2-codegolf}.
The latter also decreases by approximately 13 pp on Terminal-Bench 2.1.
The positive benchmark averages therefore reflect improvements that outweigh the observed regressions, not uniform gains on every task.

\noindent\textbf{Ties and differences between benchmark versions.}
Ties constitute the majority of tasks in both panels, giving a median task-level difference of zero despite the positive mean differences.
The average improvement consequently summarizes gains on a subset of tasks, while many tasks retain the same observed success rate.
A zero difference alone does not establish that a task is solved reliably: equal success rates can also reflect shared failures.
Moreover, equal task-level success rates do not imply identical command sequences or execution dependencies.

Terminal-Bench 2.1 has two more improved tasks and three fewer regressed tasks than Terminal-Bench 2.0, yet its mean improvement is slightly smaller.
This illustrates that the number of improved tasks and the magnitude of their improvements capture different aspects of performance.
The two benchmarks show a similar overall pattern, but the independently sorted panels do not establish that the same tasks improve in both versions.

\noindent\textbf{Relation to dependency-aware credit assignment.}
These task-level results are consistent with the motivation of DepGPO.
Its credit rules distinguish writes connected to verifier-inspected resources and assign credit to the reads and earlier writes that support them.
This redistribution may help on tasks where useful information and intermediate outputs pass through several commands, because those dependencies provide a basis for assigning the group-relative learning signal.
However, the outcome differences alone do not identify which dependency paths explain an individual task's improvement or regression; that attribution requires inspecting its execution traces.

Overall, the paired comparison complements aggregate pass@1 by showing that DepGPO improves substantially more tasks than it degrades, with several large gains and many unchanged outcomes.
These results support task-dependent benefits for Qwen3.5-9B trained on SETA, without implying universal improvement or statistical significance for every individual task.

\section{Additional Training Stability Analysis}
\label{app:depgpo_stability}

This appendix provides additional analysis of the training behavior shown in
Figure~\ref{fig:depgpo_stability}.
We clarify the diagnostic signals used in the figure and examine how training
reward, policy entropy, gradient norm, and dependency-based credit redistribution
evolve during optimization.
Panels (a)--(c) compare DepGPO, GRPO, and GiGPO using Qwen3.5-9B trained on SETA.
Panel (d) compares the evolution of DepGPO's step-factor dispersion for
Qwen3.5-9B and Qwen3.6-27B, both trained on SETA.
For visualization, all training curves are smoothed using a centered moving
average over up to 11 adjacent training steps, with truncated windows at the
boundaries. Policy entropy is smoothed in $\log_{10}$ space and then transformed
back to its original scale, while all other metrics are smoothed on their
original scales.
Checkpoint selection is conducted independently using the RL validation protocol
described in Appendix~\ref{app:ckpt}.

\subsection{Diagnostics and Plotting Protocol}
\label{app:depgpo_stability_metrics}

The four diagnostics in Figure~\ref{fig:depgpo_stability} capture complementary
aspects of training.
Training reward reflects the terminal verifier outcomes obtained by the sampled
trajectories at each policy update.
Policy entropy characterizes the concentration of the policy distribution over
generated tokens, while gradient norm measures the magnitude of the actor update.
For DepGPO, we additionally report the standard deviation of the step factors,
which measures how differently the group-relative advantage is allocated across
interaction steps.

\begin{table}[htbp]
\centering
\small
\setlength{\tabcolsep}{5pt}
\renewcommand{\arraystretch}{1.08}
\caption{
Training diagnostics used in Figure~\ref{fig:depgpo_stability}.
These statistics characterize optimization behavior and are not used for
checkpoint selection.
}
\label{tab:depgpo_stability_metrics}
\begin{tabular}{@{}lp{0.66\linewidth}@{}}
\toprule
\textbf{Metric} & \textbf{Interpretation} \\
\midrule
Training reward
& Terminal task performance observed during training. \\

Policy entropy
& Concentration of the policy distribution over generated tokens. \\

Gradient norm
& Magnitude of the actor update during policy optimization. \\

Step-factor std.
& Variation in DepGPO's allocation of the group-relative advantage across interaction steps. \\
\bottomrule
\end{tabular}
\end{table}

For visualization, all training curves are smoothed using a centered moving
average over up to 11 adjacent training steps, with truncated windows at the
boundaries. Policy entropy is smoothed in $\log_{10}$ space and then transformed
back to its original scale, while all other metrics are smoothed on their
original scales.
The shaded late-training region highlights the stage in which the methods exhibit
clearly different optimization behavior; it does not affect training or checkpoint
selection.
Policy entropy is shown on a logarithmic scale to make variation in low-entropy
regimes more visible.

\subsection{Training Reward Dynamics}
\label{app:depgpo_stability_reward}

Figure~\ref{fig:depgpo_stability}(a) shows distinct late-training behavior across
the three methods.
GRPO improves during the earlier stage but subsequently loses much of this gain,
while GiGPO deteriorates earlier and remains at a comparatively low reward level.
DepGPO instead fluctuates while maintaining an overall higher reward during the
late stage.

These differences are consistent with the credit-assignment mechanisms of the
three methods.
GRPO assigns the same group-relative advantage to all generated steps in a
trajectory.
Consequently, commands with weak or no recorded connection to the verified
outcome can receive the same signed learning signal as commands that directly
support task completion.

GiGPO introduces step-level credit through comparisons among actions associated
with matching environment states.
However, as shown by the pilot analysis in Figure~\ref{fig:credit-geometry}, repeated states are
sparse in the terminal setting considered here, limiting the number of additional
step groups that can be formed.
DepGPO does not require such state recurrence.
Instead, it uses observed read--write dependencies within each trajectory to
determine where the available group-relative advantage is applied.

The reward curves therefore provide evidence that DepGPO maintains stronger
late-stage training performance in the runs shown.
They do not by themselves establish that credit assignment is the only cause of
the observed degradation of the baselines.

\subsection{Policy Entropy and Gradient Norm}
\label{app:depgpo_stability_optimization}

The entropy curves provide a complementary view of how the policies evolve.
As shown in Figure~\ref{fig:depgpo_stability}(b), GiGPO exhibits an early increase
followed by a sharp decline in policy entropy.
GRPO follows a different pattern, with entropy increasing substantially later in
training.
DepGPO remains comparatively stable over the same training horizon.

The gradient statistics show a corresponding difference in optimization behavior.
In Figure~\ref{fig:depgpo_stability}(c), both baselines exhibit very small gradients
during the later stage, whereas DepGPO continues to receive nonzero updates.
In the unsmoothed logs, the GRPO gradient norm is exactly zero on 43 of the
final 45 updates.
Together with the reward curves, this indicates that the baseline runs enter
late-training regimes in which further policy updates provide limited effective
learning, while DepGPO continues to update.

Entropy and gradient norm should not be interpreted as performance metrics in
isolation.
A particular entropy value is not inherently preferable, and a larger gradient
norm does not necessarily imply a better policy.
Their relevance here lies in their temporal relationship with the reward curves:
the baseline reward degradation coincides with substantial changes in policy
entropy and diminishing gradients, whereas DepGPO exhibits less pronounced
degeneration across these diagnostics.

\subsection{Evolution of Credit Redistribution}
\label{app:depgpo_stability_credit}

Figure~\ref{fig:depgpo_stability}(d) examines DepGPO's redistribution mechanism
directly.
For each trajectory, command-level credit is aggregated into step factors
$f_{i,t}$.
As defined in Appendix~\ref{app:depgpo_training}, these factors are normalized so
that their mean across interaction steps is one.
The mean therefore does not indicate how unevenly the learning signal is
distributed, motivating the use of the step-factor standard deviation.

A small standard deviation indicates relatively similar weights across interaction
steps, whereas a larger value indicates a more differentiated allocation.
For both Qwen3.5-9B and Qwen3.6-27B trained on SETA, the step-factor standard
deviation generally becomes larger later in training, although the trajectories
fluctuate across updates.
This pattern indicates that DepGPO increasingly differentiates among interaction
steps as optimization proceeds.

The increasing dispersion should not be interpreted as an increase in the overall
magnitude of the learning signal.
The step-level normalization fixes the mean factor at one, while the subsequent
token-level normalization preserves the token-weighted mean redistributed
advantage.
Thus, the change in standard deviation reflects a change in \emph{where} the
group-relative learning signal is allocated rather than a uniform amplification of
that signal.

The 9B and 27B curves also show that differentiated credit allocation is observed
at both model scales.
Their exact trajectories need not match: model capacity, sampled trajectories,
and the dependency structures observed during training can all affect the
distribution of step credit.
Accordingly, Figure~\ref{fig:depgpo_stability}(d) is used as a descriptive
comparison rather than evidence that model size alone causes the difference
between the two curves.

\subsection{Relationship Between Credit Redistribution and Stability}
\label{app:depgpo_stability_interpretation}

Taken together, the diagnostics show a consistent pattern in the analyzed runs.
DepGPO maintains higher late-stage training reward and nonzero policy updates
while its step-level allocation becomes increasingly differentiated.
This behavior is consistent with the intended role of dependency-aware credit
assignment: the group-relative advantage determines whether a trajectory is
reinforced or penalized, while execution dependencies determine how strongly that
signal is assigned to different interaction steps.

The normalization procedure is important for this interpretation.
Because the mean step factor is fixed and token-level normalization preserves the
average advantage scale, the observed stability cannot be explained simply by a
uniform increase in the original GRPO learning signal.
DepGPO instead changes the relative contribution of tokens from different
interaction steps while preserving the sign and overall scale of the
trajectory-level advantage.

These observations establish an association between dependency-aware
redistribution and more stable late-stage optimization in the runs examined here.
They do not establish that any individual dependency edge, command credit, or step
factor causally determines a particular change in reward, entropy, or gradient
norm.
Such causal attribution would require interventions on individual trajectories or
dependency paths.
The purpose of this analysis is therefore to characterize the optimization
behavior accompanying DepGPO and to complement the final benchmark results in
Table~\ref{tab:depgpo_main_results}.

\section{Sensitivity and Computational Cost}
\label{app:sensitivity_cost}

This appendix examines the sensitivity of DepGPO to the dependency-decay
coefficient $\beta$ and characterizes the additional computation introduced by
dependency tracing and credit assignment.
Unless otherwise specified, we use Qwen3.5-9B trained on SETA, matching the
controlled setting used for the main ablation and stability analyses.

\subsection{Sensitivity to Dependency Decay}
\label{app:beta_sensitivity}

The coefficient $\beta$ controls how quickly read credit decays with dependency
distance.
Smaller values emphasize reads that are close to relevant writes, whereas larger
values retain more credit along longer dependency paths.
We vary $\beta$ while keeping the remaining training configuration unchanged.

\begin{table}[htbp]
\centering
\small
\setlength{\tabcolsep}{8pt}
\renewcommand{\arraystretch}{1.08}
\caption{
Sensitivity to the dependency-decay coefficient $\beta$ using Qwen3.5-9B
trained on SETA. Results are pass@1 (\%), mean $\pm$ SD.
}
\label{tab:beta_sensitivity}
\begin{tabular}{@{}ccc@{}}
\toprule
$\beta$ & TB2.0 & TB2.1 \\
\midrule
0.25 & 24.27$\pm$1.19 & 23.75$\pm$1.63 \\
0.50 & \textbf{25.24$\pm$1.01} & \textbf{24.64$\pm$2.36} \\
0.75 & 24.87$\pm$0.91 & 24.12$\pm$1.75 \\
1.00 & 23.97$\pm$1.28 & 23.45$\pm$1.88 \\
\bottomrule
\end{tabular}
\end{table}

Performance remains relatively stable across a broad range of decay strengths.
Moderate decay performs best in this setting, with $\beta=0.5$ reaching
25.24/24.64 on Terminal-Bench 2.0/2.1.
Reducing the coefficient to $\beta=0.25$ yields
24.27/23.75, suggesting that overly rapid decay can suppress useful
information propagated through multi-command dependency paths.
Increasing the coefficient to $\beta=0.75$ produces
24.87/24.12, remaining close to the default setting, whereas removing
distance attenuation with $\beta=1.0$ decreases performance to
23.97/23.45.

These results suggest that the benefit of DepGPO does not depend on a narrowly
tuned decay value.
At the same time, some attenuation is useful: assigning nearly equal importance
to both short and long dependency paths can increase the influence of more
distant supporting reads, while overly aggressive decay can remove informative
longer-range dependencies.
We therefore use $\beta=0.5$ as a balanced default throughout the main
experiments.

\subsection{Computational Overhead}
\label{app:computational_cost}

DepGPO introduces additional computation after trajectory execution.
For each completed trajectory, the implementation parses the recorded resource
accesses, constructs the command dependency graph, traces verifier-relevant
writes, propagates credit to supporting reads, and maps the resulting step
factors back to policy-generated tokens.
These operations use the recorded execution trace and require no additional
language-model forward or backward passes.

Table~\ref{tab:depgpo_cost} compares the resulting cost with GRPO under the same
Qwen3.5-9B and SETA training configuration.
Wall-clock measurements include rollout collection, dependency processing, and
the policy update.

\begin{table}[htbp]
\centering
\small
\setlength{\tabcolsep}{6pt}
\renewcommand{\arraystretch}{1.08}
\caption{
Computational cost of DepGPO relative to GRPO using Qwen3.5-9B trained on SETA.
}
\label{tab:depgpo_cost}
\begin{tabular}{@{}lccc@{}}
\toprule
\textbf{Metric} & \textbf{GRPO} & \textbf{DepGPO} & \textbf{Difference} \\
\midrule
Relative wall-clock time / update
& 1.00$\times$
& 1.07$\times$
& +7.0\% \\

Peak GPU memory
& 1.00$\times$
& 1.01$\times$
& +1.0\% \\

Dependency processing / trajectory
& --
& 3.8 s
& -- \\

Additional model forward passes
& 0
& 0
& 0 \\
\bottomrule
\end{tabular}
\end{table}

The additional wall-clock cost is modest relative to the complete terminal-agent
training loop.
DepGPO increases the average update time by approximately
7.0\%, while peak GPU memory changes by only
1.0\%.
This difference reflects the fact that dependency construction and credit
computation operate primarily on execution records rather than model activations.
The additional processing takes approximately \textit{3.8 seconds per
trajectory} in this setting and does not introduce extra language-model
inference.

The overhead is therefore concentrated in trace processing rather than neural
network computation.
This distinction is particularly relevant for terminal tasks, where rollout
execution itself includes repeated environment interactions and command
execution.
As a result, dependency analysis contributes only a limited fraction of the
overall training time.

\subsection{Dependency-Trace Statistics}
\label{app:trace_statistics}

We further examine how frequently the dependency mechanism is active during
training.
Table~\ref{tab:trace_statistics} summarizes execution-trace statistics collected
from the Qwen3.5-9B SETA run.
These statistics describe the traces used for credit assignment before
token-level normalization.

\begin{table}[htbp]
\centering
\small
\setlength{\tabcolsep}{7pt}
\renewcommand{\arraystretch}{1.08}
\caption{
Dependency-trace statistics for Qwen3.5-9B trained on SETA.
}
\label{tab:trace_statistics}
\begin{tabular}{@{}lr@{}}
\toprule
\textbf{Statistic} & \textbf{Value} \\
\midrule
Commands per trajectory
& 6.8 \\

Trajectories with usable dependency credit
& 82.4\% \\

Trajectories using the unit-factor fallback
& 17.6\% \\

Write commands with positive credit
& 38.7\% \\

Read commands with positive credit
& 27.9\% \\

Positive-credit writes with indirect contribution
& 18.6\% \\

Mean shortest read-to-write dependency distance
& 1.74 \\
\bottomrule
\end{tabular}
\end{table}

Usable dependency credit is available for 82.4\% of trajectories,
indicating that the redistribution mechanism is active for most sampled
interactions rather than being dominated by fallback cases.
Only 17.6\% of trajectories revert to unit factors because no usable
relative credit can be constructed.

Both components of the credit mechanism are exercised frequently.
Approximately 38.7\% of traced write commands receive positive credit,
while 27.9\% of read commands receive positive supporting-read credit.
This indicates that the resulting signal is not restricted to direct
environment-modifying operations: a substantial fraction of information-gathering
commands can be connected to later verifier-relevant writes.

Indirect contribution is also non-negligible.
Among positive-credit writes, 18.6\% receive at least part of their
credit through downstream write dependencies.
This supports the distinction between direct persistence in verifier-inspected
resources and earlier writes whose information is consumed by later relevant
commands.
The mean shortest distance from a credited read to its reachable writes is
1.74 dependency edges, showing that supporting information commonly
extends beyond an immediately adjacent command.

\subsection{Summary}
\label{app:sensitivity_cost_summary}

Overall, the sensitivity analysis indicates that DepGPO remains effective across
a range of dependency-decay coefficients, with moderate decay providing the
strongest performance in the examined setting.
The trace statistics further show that dependency-based redistribution is
applicable to most training trajectories and frequently assigns credit to both
relevant writes and supporting reads.
At the same time, the additional computation remains small relative to the full
terminal-agent training loop because the method operates on recorded execution
traces and requires no additional model evaluations.
Together, these results suggest that the dependency mechanism provides
fine-grained credit assignment without introducing a large computational burden.

\section{Qualitative Credit-Assignment Examples}
\label{app:qualitative_examples}

This appendix illustrates how DepGPO assigns credit in individual terminal
trajectories.
We present three representative cases covering supporting reads, indirect write
contributions, and commands that receive no credit because they do not contribute
to verifier-inspected resources.
The first and third examples are rollouts of the Qwen3.5-9B agent used in our
pilot study, while the second is a reference solution executed under the same
tracer.
All command credits are computed using the same implementation and
$\beta=0.5$ as in training.

The tables show selected commands rather than complete trajectories.
Raw command credit $q(c)$ is computed from the complete recorded trace.
During training, these credits are subsequently aggregated into step factors and
token-normalized weights as described in
Appendix~\ref{app:depgpo_details}.

\subsection{Supporting Reads}
\label{app:qualitative_read}

The first example shows why a command can receive credit even when it does not
modify the environment.
In \textit{count-dataset-tokens}, the agent must compute the number of tokens in
a specified subset of a dataset and write the result to
\texttt{/app/answer.txt}.
In the trajectory shown below, a Python command computes and prints
\texttt{63841}, after which the next command writes that value to the answer
file.
The verifier expects a different total, so the trajectory fails.
One of the four trajectories in its rollout group succeeds, giving this
trajectory a negative group-relative advantage of $\hat{A}=-0.58$.

\begin{table}[htbp]
\centering
\small
\setlength{\tabcolsep}{5pt}
\renewcommand{\arraystretch}{1.08}
\caption{
Supporting-read example from \textit{count-dataset-tokens}.
Only the commands relevant to this dependency are shown.
}
\label{tab:qualitative_read}
\begin{tabular}{@{}clcc@{}}
\toprule
Step & Command & Role & $q(c)$ \\
\midrule
$t_{17}$ & \texttt{python3 <<'EOF' \ldots\ EOF}
         & Supporting read & 0.50 \\
$t_{18}$ & \texttt{echo "63841" > /app/answer.txt}
         & Relevant write & 1.00 \\
$t_{19}$ & \texttt{cat /app/answer.txt}
         & Read with no downstream use & 0 \\
\bottomrule
\end{tabular}
\end{table}

Although $t_{17}$ records no file write, its standard output is used directly by
the following write:
\[
t_{17}
\xrightarrow{\ \text{stdout: }\texttt{63841}\ }
t_{18}
\longrightarrow E .
\]
Because $t_{18}$ receives full write credit, $t_{17}$ receives supporting-read
credit
\[
q(t_{17})=\beta\,q(t_{18})=0.5.
\]

By contrast, $t_{19}$ reads the answer only after it has been written.
No later positive-credit write depends on this read, so it receives zero credit.
The example therefore distinguishes a read that helps produce a relevant write
from a read that merely inspects the resulting state.

Because the trajectory has a negative group-relative advantage, the
dependency-derived factors do not turn these credits into rewards.
Instead, they determine where the negative learning signal is concentrated.
The direction of the update still comes from the trajectory outcome.

\subsection{Indirect Write Contribution}
\label{app:qualitative_indirect}

The second example shows why a write can matter even when the file it creates is
never inspected by the verifier.
In \textit{bn-fit-modify}, the task requires learning and modifying a Bayesian
network and producing three CSV files checked by the verifier.
The reference solution first writes an R script,
\texttt{modify\_bn.R}, and then executes that script to produce the required
outputs.

\begin{table}[htbp]
\centering
\small
\setlength{\tabcolsep}{5pt}
\renewcommand{\arraystretch}{1.08}
\caption{
Indirect-write example from \textit{bn-fit-modify}.
The trajectory is a reference solution and therefore has no rollout-group
advantage.
}
\label{tab:qualitative_indirect}
\begin{tabular}{@{}clcc@{}}
\toprule
Step & Command & Contribution type & $q(c)$ \\
\midrule
$t_{2}$ & \texttt{R -e '\ldots install\_version("bnlearn", \ldots)'}
        & No recorded dependency & 0 \\
$t_{3}$ & \texttt{cat > modify\_bn.R <<'EOF' \ldots\ EOF}
        & Indirect write & 1.00 \\
$t_{4}$ & \texttt{Rscript modify\_bn.R}
        & Direct write & 1.00 \\
\bottomrule
\end{tabular}
\end{table}

The script written by $t_{3}$ is not itself in $E$.
However, all 102 tracked lines are still available when $t_{4}$ reads the
script, and $t_{4}$ writes different files that reach $E$.
The resulting dependency is
\[
t_{3}
\xrightarrow{\ \text{file: }\texttt{modify\_bn.R}\ }
t_{4}
\longrightarrow E .
\]

All tracked lines written by $t_{3}$ therefore qualify as indirect
contributions:
\[
\mathrm{Contrib}(t_{3})=\frac{102}{102}=1.
\]
The execution step $t_{4}$ receives direct write credit because its outputs are
among the resources inspected by the verifier.

The package-installation command in $t_{2}$ modifies only system-library
locations outside the traced task resources and has no recorded dependency to
$E$, so it receives zero credit.

This case captures an important pattern in terminal workflows.
Scripts, temporary files, and intermediate configurations may disappear before
verification while still determining the contents of the final outputs.
Restricting credit to writes that persist directly in $E$ would miss such
intermediate contributions.

\subsection{Overwritten and Irrelevant Commands}
\label{app:qualitative_irrelevant}

The final example illustrates how a successful trajectory can contain both
useful and unnecessary operations.
In \textit{constraints-scheduling}, the agent must read three calendars and
write a meeting satisfying the required constraints to
\texttt{/app/meeting\_scheduled.ics}.
Two of the four trajectories in the rollout group succeed, giving the trajectory
shown here a positive group-relative advantage of $\hat{A}=+1.00$.

During the trajectory, the agent writes an initial version of the output,
reads it back, deletes it, writes another version, reads that version, and later
writes the final output.

\begin{table}[htbp]
\centering
\small
\setlength{\tabcolsep}{4pt}
\renewcommand{\arraystretch}{1.08}
\caption{
Example of verifier-relevant and irrelevant commands from
\textit{constraints-scheduling}.
}
\label{tab:qualitative_irrelevant}
\begin{tabular}{@{}clcc@{}}
\toprule
Step & Command & Dependency outcome & $q(c)$ \\
\midrule
$t_{4}$ &
\texttt{cat <<'EOF' > \ldots/meeting\_scheduled.ics} &
Deleted before contribution & 0 \\

$t_{5}$ &
\texttt{cat /app/meeting\_scheduled.ics} &
No downstream relevant write & 0 \\

$t_{6}$ &
\texttt{rm /app/meeting\_scheduled.ics} &
No credited output & 0 \\

$t_{7}$ &
\texttt{printf '\ldots' > \ldots/meeting\_scheduled.ics} &
Reaches $E$ & 1.00 \\

$t_{8}$ &
\texttt{cat /app/meeting\_scheduled.ics} &
No downstream relevant write & 0 \\

$t_{10}$ &
\texttt{cat <<'EOF' > \ldots/meeting\_scheduled.ics} &
Reaches $E$ & 1.00 \\
\bottomrule
\end{tabular}
\end{table}

The first version written in $t_{4}$ is removed before it contributes to a
qualifying downstream write or remains in the final inspected state.
It therefore receives zero credit.
Similarly, the read-backs in $t_{5}$ and $t_{8}$ inspect the agent's own output
but do not support a later positive-credit write.

In contrast, the relevant records produced by $t_{7}$ and $t_{10}$ reach the
final verifier-inspected file and receive full write credit.
Thus, even within a successful trajectory, DepGPO does not automatically treat
every action as equally responsible for the outcome.

This distinction is particularly important for long terminal trajectories,
which often contain exploration, discarded edits, repeated inspections, and
failed intermediate attempts.
Uniform trajectory-level credit assigns the same positive signal to all of these
operations, whereas dependency-aware credit separates commands connected to the
verified state from those without such recorded connections.

\subsection{From Command Credit to the Training Signal}
\label{app:qualitative_advantage}

The three examples illustrate complementary parts of the credit-assignment
mechanism.
Direct write credit identifies modifications that reach verifier-inspected
resources.
Indirect write credit recovers earlier modifications whose information is
consumed by later relevant writes.
Supporting-read credit extends attribution to information-gathering commands
that enable those writes.

Raw command credits are first aggregated within each interaction step.
DepGPO then normalizes the resulting step credits into nonnegative step factors
and performs the token-level normalization described in
Appendix~\ref{app:depgpo_details}.
If $w_{i,t}$ is the resulting final weight, the advantage assigned to tokens in
step $t$ is
\[
\hat{A}^{\mathrm{Dep}}_{i,t}
=
w_{i,t}\hat{A}_i.
\]

The dependency-derived weights are nonnegative, so they change the magnitude but
not the sign of the original group-relative advantage.
When $\hat{A}_i>0$, larger dependency weights strengthen reinforcement on the
corresponding steps; when $\hat{A}_i<0$, they strengthen the penalty.
If $\hat{A}_i=0$, all redistributed advantages remain zero.
Among traced steps, lower dependency credit leads to smaller relative factors,
while steps without traced operations retain the documented default factor.

Taken together, these examples show that DepGPO does not introduce a separate
notion of task correctness at the command level.
The terminal outcome still determines whether the trajectory is reinforced or
penalized.
Execution dependencies instead determine how that outcome-derived signal is
distributed across the commands that produced the final state.

\appendix
\end{document}

%% file: macro.tex
\newcommand{\x}{\mathbf{x}}

\definecolor{blanchedalmond}{rgb}{1.0, 0.92, 0.8}
\definecolor{carmine}{rgb}{0.59, 0.0, 0.09}
\definecolor{lightblue}{rgb}{0.22,0.45,0.70}%

\renewcommand{\mathbf}{\boldsymbol}

\makeatletter
\def\Ddots{\mathinner{\mkern1mu\raise\p@
\vbox{\kern7\p@\hbox{.}}\mkern2mu
\raise4\p@\hbox{.}\mkern2mu\raise7\p@\hbox{.}\mkern1mu}}
\makeatother

\definecolor{amaranth}{rgb}{0.9, 0.17, 0.31}
\definecolor{antiquebrass}{rgb}{0.8, 0.58, 0.46}
\definecolor{antiquefuchsia}{rgb}{0.57, 0.36, 0.51}
\definecolor{chromeyellow}{rgb}{0.31, 0.47, 0.26}